\documentclass[preprint, authoryear]{elsarticle}

\usepackage{amsmath}
\usepackage{amssymb}
\usepackage{mathtools}
\usepackage{amsthm}
\usepackage{inputenc}
\usepackage{graphicx}
\usepackage{booktabs} 
\usepackage{hyperref}
\usepackage{xcolor}
\usepackage{multirow}
\usepackage{array}
\usepackage{pifont}
\usepackage{indentfirst}
\usepackage{subcaption}
\usepackage{caption}

\newcommand{\dx}{\, \mathrm{d}x}
\graphicspath{{img_}}

\newtheorem{thm}{Theorem}
\newtheorem{lem}{Lemma}
\newtheorem{prop}{Proposition}
\newtheorem{cor}{Corollary}
\newdefinition{rmk}{Remark}
\newproof{pf}{\textbf{Proof}}

\newcolumntype{C}[1]{>{\centering\arraybackslash}m{#1}}

\DeclareMathOperator*{\argmin}{argmin}
\newcommand{\name}{$\Psi$-GNN}

\begin{document}

\begin{frontmatter}

\title{An Implicit GNN Solver for Poisson-like Problems}

\author[1,2]{Matthieu Nastorg\corref{cor1}} 
\ead{matthieu.nastorg@inria.fr} 

\author[1,3]{Michele-Alessandro Bucci}
\author[2]{Thibault Faney}
\author[2]{Jean-Marc Gratien}
\author[1]{Guillaume Charpiat}
\author[1]{Marc Schoenauer}

\cortext[cor1]{Corresponding author}

\affiliation[1]{organization={Université Paris-Saclay, CNRS, Inria, Laboratoire interdisciplinaire des sciences du numérique},
postcode={91405},
city={Orsay},
country={France}}
\affiliation[2]{organization={IFP Energies nouvelles},
postcode={92852},
city={Rueil-Malmaison},
country={France}}
\affiliation[3]{organization={Safran Tech, Digital Sciences \& Technologies Department},
postcode={78114},
city={Châteaufort},
country={France}}

\begin{abstract}
This paper presents \name{}, a novel Graph Neural Network (GNN) approach for solving the ubiquitous Poisson PDE problems on general unstructured meshes with mixed boundary conditions. By leveraging the Implicit Layer Theory, \name{} models an ``infinitely'' deep network, thus avoiding the empirical tuning of the number of required Message Passing layers to attain the solution. Its original architecture explicitly takes into account the boundary conditions, a critical pre-requisite for physical applications, and is able to adapt to any initially provided solution. \name{} is trained using a physics-informed loss, and the training process is stable by design. Furthermore, the consistency of the approach is theoretically proven, and its flexibility and generalization efficiency are experimentally demonstrated: the same learned model can accurately handle unstructured meshes of various sizes, as well as different boundary conditions. To the best of our knowledge, \name{} is the first physics-informed GNN-based method that can handle various unstructured domains, boundary conditions and initial solutions while also providing convergence guarantees.
\end{abstract}

\end{frontmatter}

\section{Introduction}
\label{introduction}

Partial differential equations (PDEs) are highly detailed mathematical models that describe complex physical or artificial processes in engineering and science \citep{intro_pdes}. Although they have been widely studied for many years, solving these equations at scale remains daunting, limited by the computational cost of resolving the smallest spatio-temporal scales \citep{num_sol_pdes}. One of the most common and important PDEs in engineering is the steady-state \textit{Poisson} equation, mathematically described as 
$ - \Delta u = f$
on a domain $\Omega \subset \mathbb{R}^n$ (usually, $n=2$ or $3$) with a given {\em forcing term} $f$, where $u$ is the solution to be sought. Besides, special constraints (often referred to as \textit{boundary conditions}) must be specified on the boundary $\partial \Omega$ of $\Omega$ \citep{intro_num_meth_var_prob} to ensure the existence and uniqueness of the solution. The Poisson equation appears in various fields such as Fluid mechanics \citep{pressure_poisson}, Gravity \citep{gravity_poisson}, Electrostatics \citep{electrostatics_poisson}, or Surface reconstruction \citep{surface_reconstruction_poisson}. It is ubiquitous and plays a central role in modern numerical solvers. Despite progresses in the High-Performance Computing (HPC) community, solving large Poisson problems is achievable only by employing robust yet tedious iterative methods \citep{iterative_method_sparse_linear_systems} and remains one of the major bottlenecks in the speedup of industrial numerical simulations. 
\\\\
More recently, data-driven methods based on deep neural networks have been reshaping the domain of numerical simulation. Neural networks can provide faster predictions, reducing the turnaround time for workflows in engineering and science \citep{latent_space_physics, solver_in_the_loop, ml_accelerated_cfd} -- see also our quick survey in Section \ref{related_work}. However, the lack of guarantees about the consistency and convergence of deep learning approaches makes it non-viable to implement these models in the design and production stage of new engineering solutions.
\\ \\
This work introduces \name{}\footnote{Poisson Solver Implicit Graph Neural Network}, an Implicit Graph Neural Network (GNN) approach that iteratively solves a Poisson problem on general unstructured meshes with mixed boundary conditions. Leveraging Implicit Layer Theory \citep{deq}, the proposed model controls, by itself, the number of Message Passing layers needed to reach the solution, yielding excellent out-of-distribution generalization to mesh sizes and shapes. The \name{}{} architecture, detailed in Section \ref{methodology}, is based on a node-oriented approach which inherently respects the boundary conditions, and an additional autoencoding process allows it to consider any initial solution and dynamically adapt to it. The method is trained end-to-end, minimizing the residual of the discretized Poisson problem, thus attempting to learn the physics of the problem. When Neumann boundary conditions are present, an additional lightly-supervised loss (MSE with ground-truth solutions) is necessary to help convergence. To ensure the stability of the model, a regularizing cost function \citep{deq_stab} is also used to constrain the spectral radius of the Machine Learning solver, thus providing strong convergence guarantees. 
In Section \ref{theory}, we provide a theoretical analysis, demonstrating that the proposed \name{} approach satisfies a property of universal approximation. Specifically, we prove that there exists a parametrization of our model that yields an optimal solution for the considered task, showcasing the consistency of our approach. Moreover, we verify its generalization ability in Section \ref{results} through experiments with various geometries and boundary conditions. We also assess the performance of our approach by comparing it with a state-of-the-art Machine Learning model and discuss its complexity.
\\\\
\textbf{Significant Contributions of this Study:}
\begin{itemize}
    \item Introduction of a novel approach that combines Graph Neural Networks and Implicit Layer Theory for effectively addressing Poisson problems with mixed boundary conditions.
    \item Development of a flexible model capable of accommodating diverse mesh sizes and shapes, as well as handling various boundary conditions and initial solutions, demonstrating excellent out-of-distribution generalization.
    \item Implementation of a robust training process that learns the physics of the equation and uses stabilization techniques to provide strong convergence guarantees.
\end{itemize}

\section{Related Work}
\label{related_work}

In the past few years, the use of Machine Learning models to predict solutions of PDEs has gained significant interest in the community, beginning in the 90s with some pioneer works of \citet{neural_algo_for_solving_diff_eq, nn_app_for_solving_pdes} and \citet{artificial_nn_for_solving_odes_and_pde}. Since then, much research has focused on building more complex neural network architectures with a larger number of parameters, taking advantage of the increasing computational power as demonstrated in \citet{modelling_dynamics_of_nonlinear_pdes, ann_for_solving_stokes_problem} or \citet{mlp_and_radial_basis_nn}. 
\\\\
\textbf{CNNs for physics simulations} 
\quad Despite these convincing advances that employed fully connected neural networks, these methods were quickly overtaken by the tremendous progress in computer vision and the rise of Convolutional Neural Networks (CNNs) \citep{cnn_lecun, cnn_advances}. For instance, in fluid mechanics, such networks have been used to solve the Navier-Stokes equations, considering solutions on rectangular grids and treating them as images \citep{cnn_stead_flow, cnn_approach_training_airfoil_performance, cfdnet, performance_and_accuracy_assessments}. A significant amount of research has focused on using CNNs to solve the Poisson equation due to its significant engineering interest. In \citet{poisson_solver_cnn}, a straightforward CNN architecture predicts the potential electric distribution in a square domain by approximating the solution of 2D and 3D Poisson problems. \citet{learning_pde_solver_convergence_guarantees} design a Machine Learning solver with a U-Net architecture \citep{u_net} to mimic multigrid methods \citep{multigrid} and provide theoretical convergence guarantees when applied to the resolution of a 2D Poisson equation. In \citet{poisson_cnn}, a convolutional neural network is trained to solve the inverse Poisson problem through supervised learning. In \citet{poisson_plasma_flow}, the authors use a physics-based loss function with a deep convolutional network to solve the Poisson equation in the context of plasma flows. These different approaches have shown promising results, providing rather accurate approximate solutions that are also computed faster than traditional solvers. However, there are limitations to their application. They can only be used with structured meshes with uniform discretization, which makes them incompatible with classical methods that rely on unstructured ``\textit{meshes}". Another issue is that the treatment of boundary conditions is often overlooked or not properly addressed. If attempts are made to consider them, it is usually done by adding an extra loss function or using an additional specialized model. Unfortunately, these approaches tend to perform poorly when dealing with new problem configurations, as they struggle to generalize effectively.
\\\\
\textbf{GNNs for physics simulations}
\quad To address these shortcomings, recent studies have focused on Graph Neural Networks (GNNs), a class of neural networks that can learn from unstructured data. Introduced in \citet{fundation_gnn}, GNNs have experienced significant growth and seen a variety of applications thanks to the development of new techniques such as graph convolution \citep{graph_convolution_networks}, edge convolution \citep{edge_conv}, graph attention networks \citep{graph_attention_networks}, or graph pooling \citep{graph_pooling} to name a few. An exhaustive survey on existing GNN architectures can be found in \citet{survey_graph_network}. With regard to physical applications, several recent works have shown the ability of GNNs to learn dynamical systems accurately. For example, in \citet{compo_object_based_learn_physical_dynamics} and \citet{graph_networks_as_learnable_physics_engines}, GNNs are used to learn the motion of discrete systems of solid particles. \citet{learning_to_simulate_complex_physics} extend this approach to learn complex physics, including fluid simulation and solid deformation, considering graph nodes as particles. In \citet{learning_mesh_based_simulation_with_graph_networks}, the authors simulate the time dynamics of complex systems based on unstructured data. \citet{multipole_graph_neural_operator} and \citet{simulating_continuum_mechanics} propose using a multi-level architecture to solve PDEs on graphs with a larger number of nodes. In \citet{physics_embedded_nn}, a GNN model is trained through supervised learning to approximate the solution of the incompressible Navier-Stokes equation while preserving the boundary conditions. Similar to CNN-oriented research, some studies have focused on using GNNs to solve the Poisson equation, starting with the work of \citet{poisson_gnn_fundation}. \citet{neural_operator} introduce a graph kernel network to approximate PDEs with a specific focus on the resolution of a 2D Poisson problem, and in \citet{machine_learning_based_solver_pressure_poisson}, a multi-level GNN architecture is trained through supervised learning to solve the Poisson Pressure equation in the context of fluid simulations.  These approaches outperform the CNN-based models due to their generalization to unstructured meshes. Nevertheless, these methods still rely solely on supervised learning, requiring computationally expensive ground truth solutions and resulting in poor performances when applied to out-of-distribution examples. Additionally, the explicit consideration of boundary conditions remains elusive, presenting a significant challenge for practical use in industrial processes.
\\\\
\textbf{The Physics-Informed approach}
\quad In parallel to these architectural advancements, another research direction focuses on a new class of Deep Learning method called \textit{Physics-Informed Neural Networks} (PINNs), which has emerged as a very promising tool for solving scientific computational problems \citep{pinns_fundation, pinns_fundation_2}. These methods are mesh-free and specifically designed to integrate the PDE residual into the training loss. To do that, PINNs leverage Automatic Differentiation \citep{automatic_differentiation} to compute the PDE's derivatives. Numerous works have used this approach to solve more complex problems, such as in fluid mechanics as demonstrated in \citet{pinns_turbulence, pinns_high_speed_flow, pinns_laminar_flow, pinns_incompressible_ns}, or \citet{pinns_review}. Therefore, the approach of training a model by minimizing the residual equation is well known and has been combined with GNNs to solve a 2D Poisson equation as in \cite{pinns_galerkin_method}. However, PINNs face challenges in generalizing to new scenarios due to the significant changes that can occur in the solution to a PDE when considering different domain shapes or boundary conditions. Moreover, effectively addressing the boundary conditions often requires the inclusion of additional loss terms, which must be carefully weighted to achieve optimal performance. Furthermore, PINNs lack interpretability. In our approach, we build upon established numerical methods by directly minimizing the residual of the discretized Poisson problem. By doing so, we inherit the advantageous properties of the chosen discretization method, i.e. the Finite Element method (FEM). As a result, our model adopts a mesh-based structure and effectively incorporates the strengths and characteristics of the FEM. Consequently, our approach demonstrates enhanced performance and an improved ability to generalize across different domains, boundary conditions, and initial solutions.
\\\\
\textbf{Deep Statistical Solvers (DSS)}
\quad Our work is closely related to \citet{deep_statistical_solver}, where a GNN is used to solve a Poisson problem with Dirichlet boundary conditions efficiently. We propose a novel GNN-based architecture that can handle mixed boundary conditions, making it extensible to CFD cases. In previous work, the number of Message Passing Neural Networks (MPNN) needed to achieve convergence was fixed and was shown to be proportional to the diameter of meshes considered in the dataset. \citet{ds_gps} further improve upon that work by introducing a Recurrent Graph architecture which significantly reduces the model size. Besides, they show that if the model is trained with a sufficient number of MPNN iterations, it tends to converge toward a fixed point. However, the number of iterations remains fixed, and the model has poor generalization capabilities to different mesh sizes. To address this issue, we propose using the Implicit Layer Theory \cite{deq} to model an infinitely deep neural network. Furthermore, we propose an architecture that employs an autoencoding process and a node-specific approach to dynamically adapt to any given initial solution and ensure that the boundary conditions are respected by design.
\\\\
Building upon the advancements made by these state-of-the-art models, our research aims to create a Machine Learning algorithm that can effectively and accurately solve a wide range of Poisson problems. In particular, we aim to enhance the existing DSS framework by proposing a novel GNN-``physics-oriented" model. \textbf{This model is specifically designed to address the challenges posed by unstructured meshes with varying sizes and shapes while ensuring the explicit incorporation of boundary conditions.}

\section{Methodology}
\label{methodology}

In this section, we detail the methodology employed in this study. We begin by stating the problem in \ref{problem_statement}. Then, in \ref{statistical_problem}, we delve deeper into the complexity of the problem by providing a graph interpretation and presenting the statistical problem at hand. In \ref{architecture}, we outline the design of our model, providing a comprehensive description of its architecture. Additionally, we discuss in \ref{stabilization} the regularization technique employed to ensure the stability of the method and provide further details on the training process in \ref{training}.

\subsection{Problem statement}
\label{problem_statement}

 Let $\Omega \subset \mathbb{R}^n$ be a bounded open domain with smooth boundary $\partial \Omega = \partial \Omega_D \cup \partial \Omega_N $. 
To ensure the existence and uniqueness of the solution, special constraints (referred to as \textit{boundary conditions}) must be specified on the boundary $\partial \Omega$ of $\Omega$ \citep{intro_num_meth_var_prob}: {\em Dirichlet conditions} assign a known value for the solution $u$ on $\partial \Omega_D$, and  {\em homogeneous Neumann boundary conditions} impose that no part of the solution $u$ is leaving the domain through $\partial \Omega_N$. More formally, let $f$ be a continuous function defined on $\Omega$, $g$ a continuous function defined on $\partial \Omega_D$ and $n$ the outward normal vector defined on $\partial \Omega$. The Poisson problem with mixed boundary conditions (i.e. Dirichlet and homogeneous Neumann boundary conditions) consists in finding a real-valued function $u$, defined on $\Omega$, that satisfies:

\begin{equation}
\left \{
\begin{array}{rcl}
-\Delta u &=& f \qquad \in \Omega \\
u &=& g \qquad \in \partial \Omega_D \\
\frac{\partial u}{\partial n} &=& 0 \qquad \in \partial \Omega_N
\end{array}
\right.
\label{poisson-equation}
\end{equation}

 Except in very specific instances, no analytical solution can be derived for the Poisson problem, and its solution must be numerically approximated: the domain $\Omega$ is first discretized into an unstructured mesh, denoted $\Omega_h$. The Poisson equation (\ref{poisson-equation}) is then spatially discretized using the Finite Element Method (FEM) \citep{finite_element_method}. The approximate solution $u_h$ is sought as a vector of values defined on all $N$ degrees of freedom of $\Omega_h$. $N$ in turn depends on the type of elements chosen (i.e., the precision order of the approximation wanted): choosing first-order Lagrange elements, $N$ matches the number of nodes in $\Omega_h$. The discretization of the variational formulation using Galerkin's method leads to solving a linear system of the form:
\begin{equation}
AU = B
\label{linear-system}
\end{equation} 
where the sparse matrix $A \in \mathbb{R}^{N\times N}$ is the discretization of the continuous Laplace operator, the vector $B \in \mathbb{R}^N$ comes from the discretization of the forcing term $f$ and of the mixed boundary conditions, and $U \in \mathbb{R}^N$ is the solution vector to be sought. Details regarding the derivation of the linear system \eqref{linear-system} can be found in \ref{appendix:fem}.
\\\\
Let $\mathcal{F}$ be a set of continuous functions on $\Omega$ and $\mathcal{G}$ a set of continuous functions on $\partial \Omega_D$. We denote as $\mathcal{P}$ a set of Poisson problems, 
% parametrized by $p \in \Sigma$, 
such that any element $E_p \in \mathcal{P}$ is described as a triplet:
$$E_p = \left( \Omega_{p}, ~f_p, ~g_p \right)$$ 
where $f_p \in \mathcal{F}$ and $g_p \in \mathcal{G}$. For all $E_p \in \mathcal{P}$, let $E_{h,p} \in \mathcal{P}_h$ denote its discretization, such that: 
$$E_{h,p} = \left(\Omega_{h,p}, ~A_p, ~B_p \right)$$ where $A_p$ and $B_p$ are defined as in \eqref{linear-system}).
\\\\
 The fundamental idea of \name{} is, considering a continuous Poisson problem $E_p \in \mathcal{P}$, to build a Machine Learning solver, parametrized by a vector $\theta$, which outputs a solution $U_p$ of its respective discretized form $E_{h,p} \in \mathcal{P}_h$: 
\begin{equation}
U_p ~ = ~ \text{\name}_\theta\left(E_{h,p}\right) = ~ \text{\name}_\theta\left(\Omega_{h,p}, ~A_p, ~B_p\right) 
\end{equation}

\subsection{Statistical problem}
\label{statistical_problem}

\begin{figure}[t!]
     \centering
     \begin{subfigure}[b]{0.45\textwidth}
         \centering
         \includegraphics[scale=0.35]{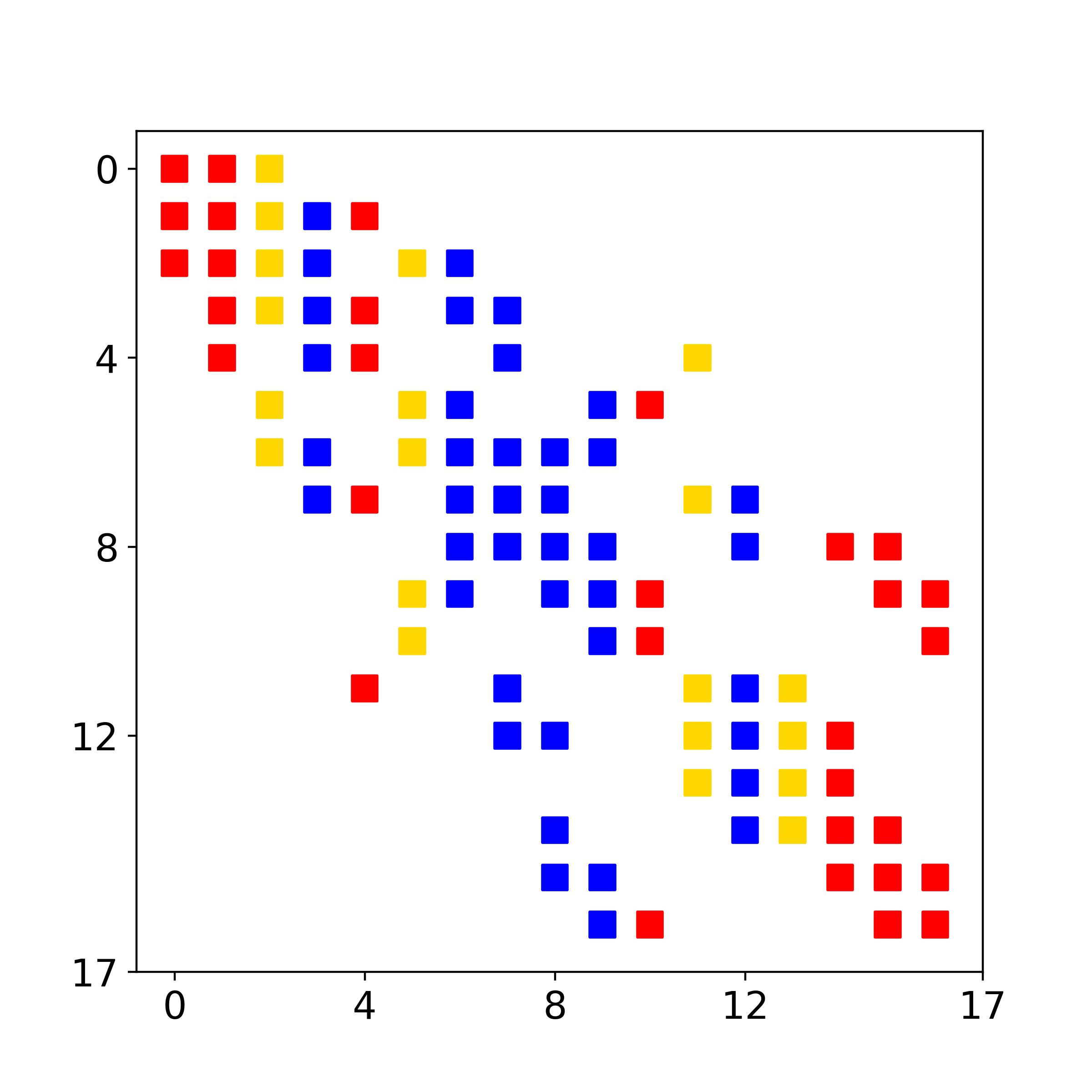}
         \caption{}
         \label{fig:sparsenobounds}
     \end{subfigure}
     \hfill
     \begin{subfigure}[b]{0.45\textwidth}
         \centering
         \includegraphics[scale=0.35]{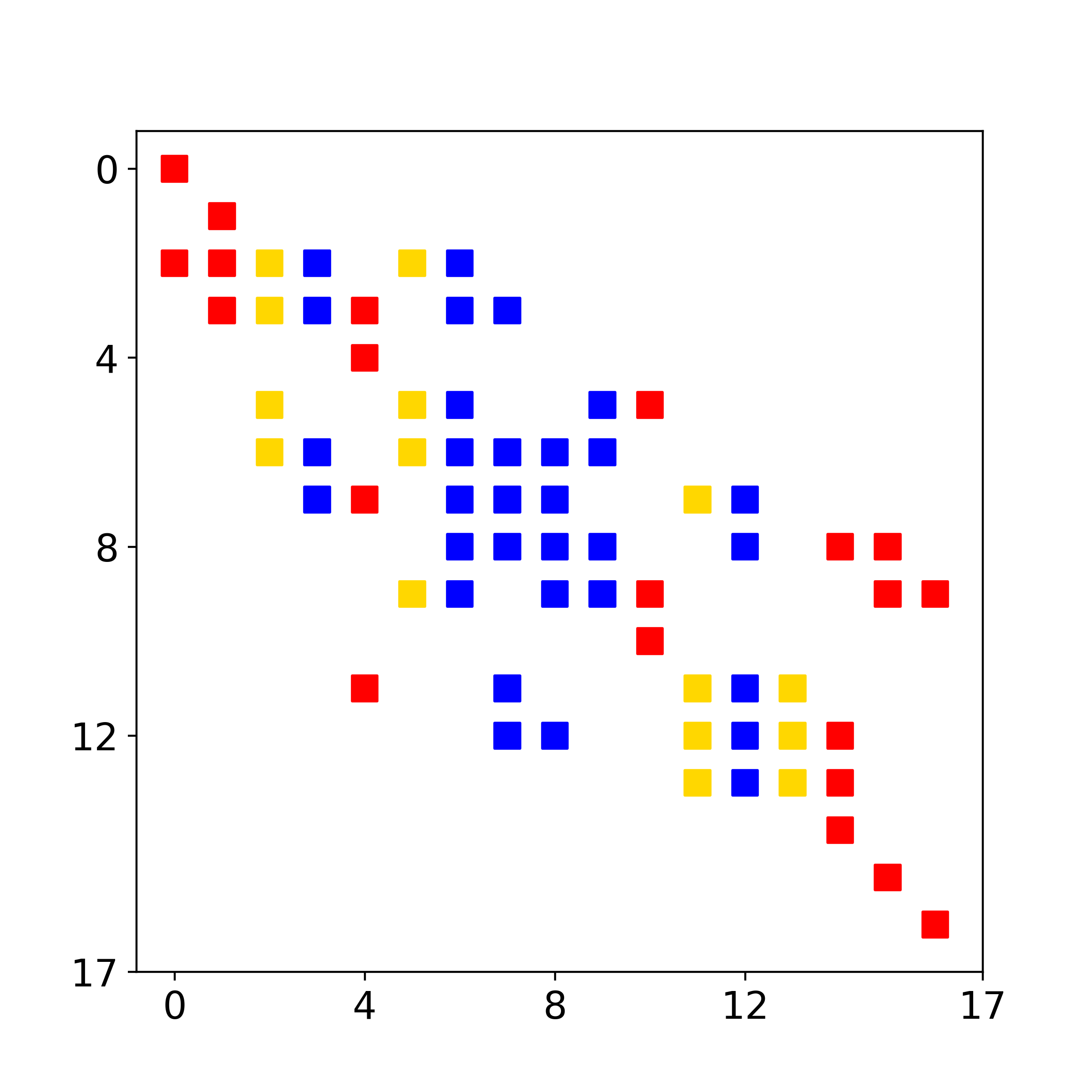}
         \caption{}
         \label{fig:sparsebounds}
     \end{subfigure}
     \hfill
     \par\bigskip\bigskip
     \begin{subfigure}[b]{0.5\textwidth}
         \centering
         \includegraphics[scale=0.07]{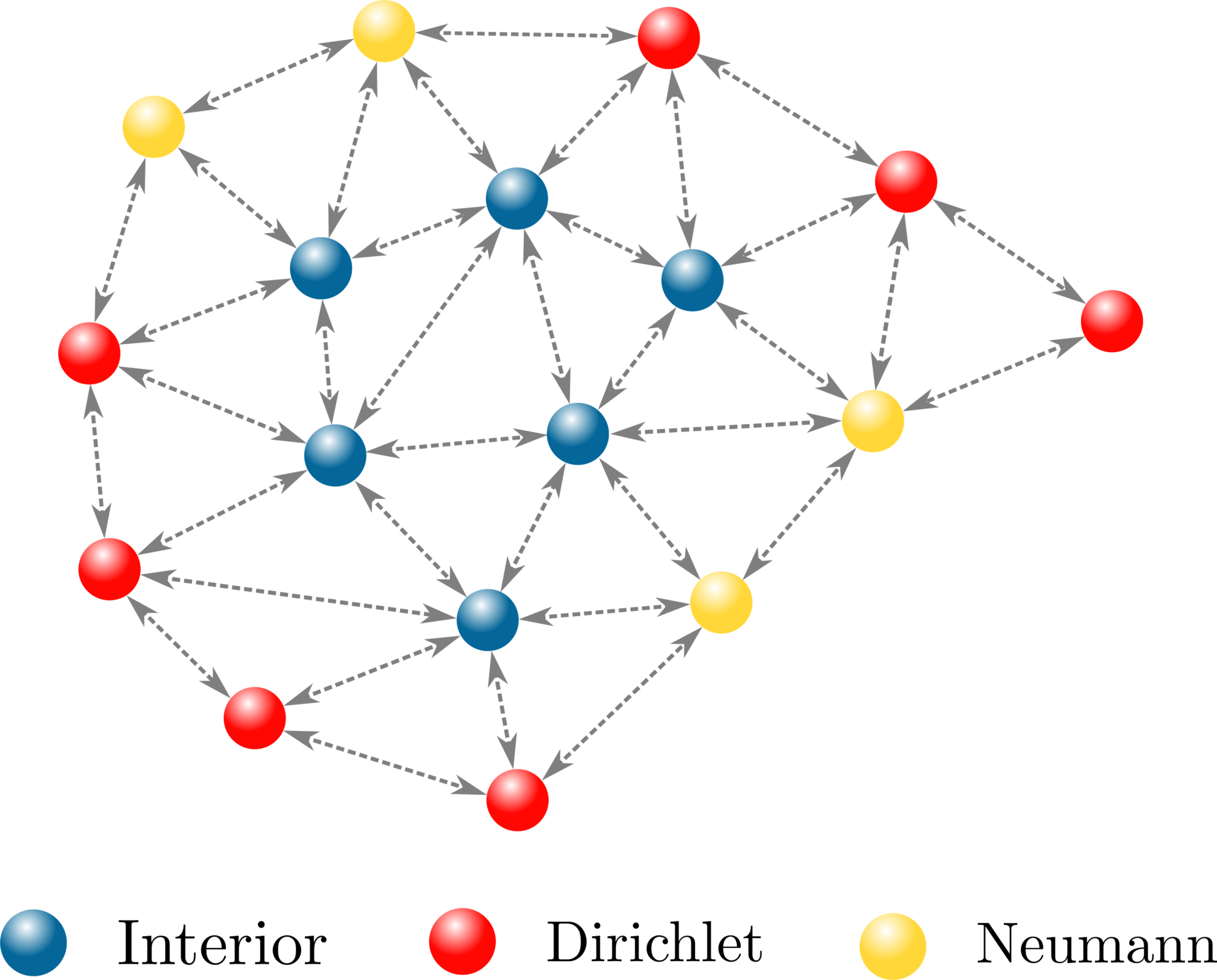}
         \caption{}
         \label{fig:graphsketchnobound}
     \end{subfigure}
     \hfill
     \begin{subfigure}[b]{0.45\textwidth}
         \centering
         \includegraphics[scale=0.07]{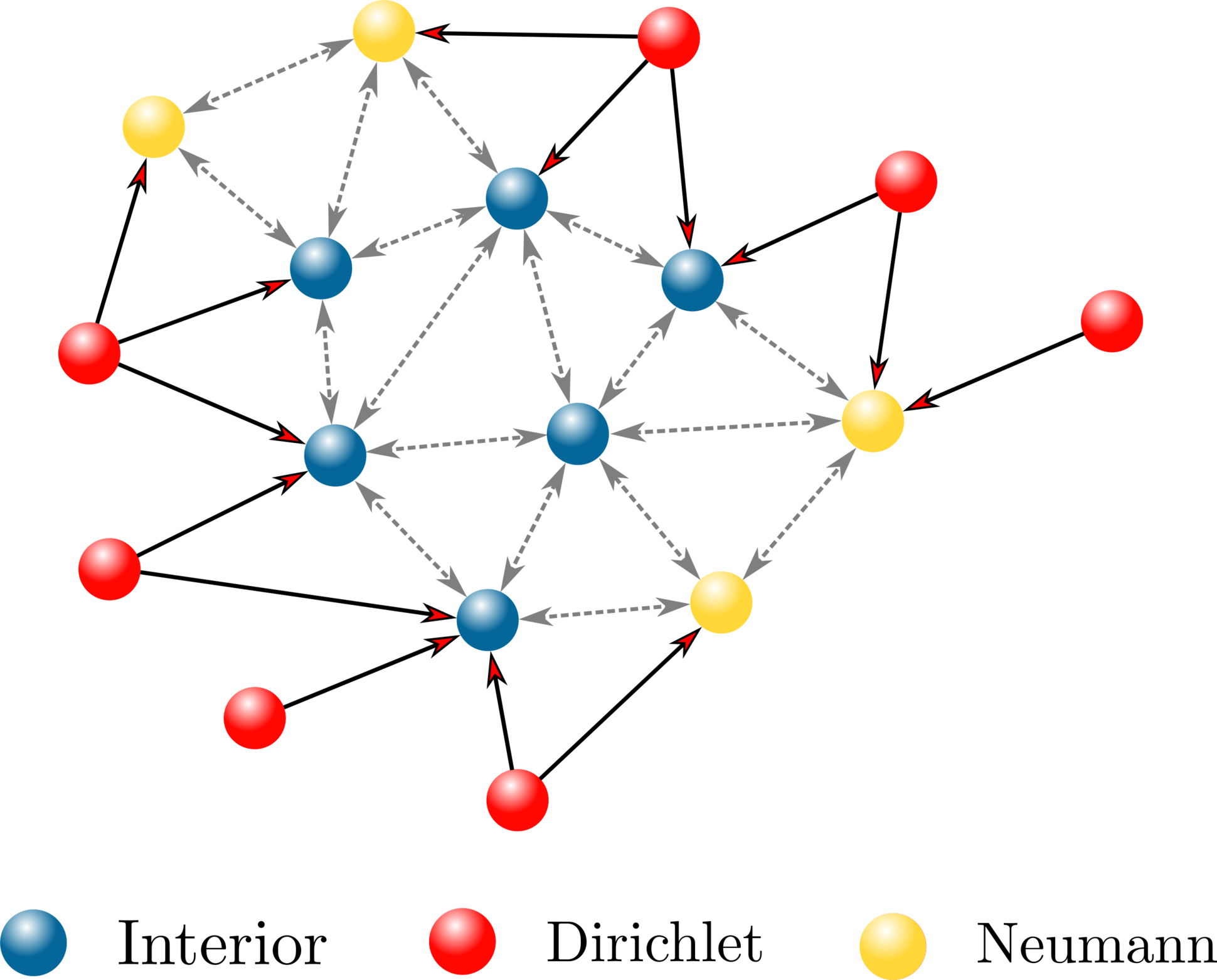}
         \caption{}
         \label{fig:graphsketchbound}
     \end{subfigure}
     \hfill
     \caption{(\ref{fig:sparsenobounds}) illustrates the sparsity pattern in matrix $A$, obtained by discretizing the Laplace operator in \eqref{poisson-equation} using FEM for a problem with 17 nodes \textit{before} applying Dirichlet boundary conditions. Using this matrix as an adjacency matrix, the induced graph is shown in (\ref{fig:graphsketchnobound}), resulting in a fully undirected graph. (\ref{fig:sparsebounds}) displays the sparsity pattern of the same matrix $A$ \textit{after} applying Dirichlet boundary conditions. The related graph is shown in (\ref{fig:graphsketchbound}), resulting in an undirected graph for Interior and Neumann nodes (blue and yellow nodes) and a directed graph for Dirichlet nodes (red nodes). For both (\ref{fig:sparsenobounds}) and (\ref{fig:sparsebounds}), Interior, Neumann and Dirichlet connections correspond to the blue, yellow and red squares, respectively.}
\end{figure}

In (\ref{linear-system}), it should be noted that the structure of matrix $A$ encodes the geometry of its corresponding mesh (i.e. $A$ can be viewed as the adjacency matrix of its corresponding mesh). Indeed, for each node, using first-order finite elements leads to the creation of local stencils, which represent local connections between mesh nodes. When using higher-order finite elements, the number of degrees of freedom does not match the number of nodes in the mesh, and the previous assumptions are no longer valid. Hence, this work exclusively focuses on the use of first-order finite elements. 
\\\\
A crucial upside of using GNNs in physics simulations is related to the right treatment of boundary conditions. In the linear system given by \eqref{linear-system}, implementing Dirichlet boundary conditions involves modifying the sparse matrix $A$ by setting the off-diagonal elements of the rows corresponding to Dirichlet nodes to $0$ and the diagonal element to $1$. Similarly, the corresponding index in vector $B$ is set to the discrete value of $g$. As a result, when solving \eqref{linear-system}, Dirichlet conditions are enforced, ensuring that $u=g$. Figure \ref{fig:sparsenobounds} displays the sparsity pattern of matrix $A$ for a problem with $17$ nodes before applying Dirichlet boundary conditions. In contrast, Figure \ref{fig:sparsebounds} illustrates the sparsity pattern of matrix $A$ for the same problem after applying Dirichlet boundary conditions. This comparison reveals that such modifications break the symmetry of the matrix $A$. As a consequence, some Interior connections (blue squares in Figure \ref{fig:sparsebounds}) and Neumann connections (yellow squares in Figure \ref{fig:sparsebounds}), linked to Dirichlet nodes no longer have symmetrical counterparts. As a result of this specific construction, the induced graph is \textit{directed} at those Dirichlet boundary nodes, sending information only to its neighbours without receiving any. Conversely, Interior and Neumann nodes induce an \textit{undirected} graph with bi-directional edges, thus exchanging information with each other. To further illustrate this, Figure \ref{fig:graphsketchnobound} displays the graph obtained using the adjacency matrix depicted in \ref{fig:sparsenobounds} (i.e. \textit{before} applying Dirichlet boundary conditions). Since \ref{fig:sparsenobounds} is symmetrical, all edges are bi-directional. On the contrary, Figure \ref{fig:graphsketchbound} illustrates the graph obtained by considering the adjacency matrix \ref{fig:sparsebounds} (i.e. \textit{after} applying Dirichlet boundary conditions), showcasing the specific directionality of the edges based on the node types. Furthermore, the end of \ref{appendix:fem} presents a concise yet illustrative example that facilitates a comprehensive mathematical understanding of why the sparse matrix $A$ induces such a graph structure. \\

\noindent A discretized Poisson problem $E_h = \left(\Omega_h, ~A, ~B \right)$ with $N$ degrees of freedom can then be interpreted as a graph problem $G = (N, ~A, ~B)$, where $N$ is the number of nodes in the graph, $A = (a_{ij})_{(i,j) \in[N]^2}$ is the weighted adjacency matrix that represents the interactions between the nodes and $B = (b_i)_{i\in[N]}$ is the local external input. Vector $U = (u_i)_{i\in[N]}$ represents the state of the graph, $u_i$ being the state of node $i$. 
\\\\
Additionally, let $\mathcal{S}$ be the set of all such graphs $G$, $\mathcal{U}$ the set of all states $U$, and $\mathcal{L}_\text{res}$ the real-valued function which computes the mean squared error (MSE) of the discretized residual equation:
\begin{align}
\mathcal{L}_\text{res}(U,~G) & = \text{MSE}\left(AU - B\right) \\
& = \displaystyle \frac{1}{N}\sum_{i \in [N]} \Big( -b_i + \sum_{j \in [N]} a_{i,j}u_j \Big) ^2
\label{loss-function}
\end{align}
Our goal is, given a graph $G$ in $\mathcal{S}$, to find an optimal state in $\mathcal{U}$ that minimizes (\ref{loss-function}). Therefore, we define a Machine Learning algorithm \name, parameterized by $\theta$, that predicts from $G$ a solution $U$ in order to solve the following statistical problem: \\\\
\textit{Given a distribution $\mathcal{D}$ on space $\mathcal{S}$ and a loss function $\mathcal{L}_\text{res}$, solve:}
\begin{equation}
\displaystyle \widehat{\theta} = \underset{{\theta}}{\text{argmin}}~\underset{G \sim \mathcal{D}}{\mathbb{E}} \left[\mathcal{L}_\text{res}\left(\text{\name}_{\theta}\left(G\right),G\right)\right]
\label{ssp-problem}
\end{equation}

\name{} is trained using the loss function described in \eqref{loss-function}, which involves computing the matrix $A$ and vector $B$ in (\ref{linear-system}), obtained from the Finite Element method. However, the model only takes as input the mesh structure (converted into a graph), distances between nodes (used as edge features), and discretized values of the forcing term $f$ and boundary function $g$ (used as node features). As a result, once the model is trained, inference only requires information about the mesh and the original problem, eliminating the need for the expensive computation of the elements in the linear system (\ref{linear-system}).

\subsection{Architecture}
\label{architecture}

\begin{figure*}[t!]
\vskip 0.2in
\begin{center}
\centerline{\includegraphics[width=\textwidth]{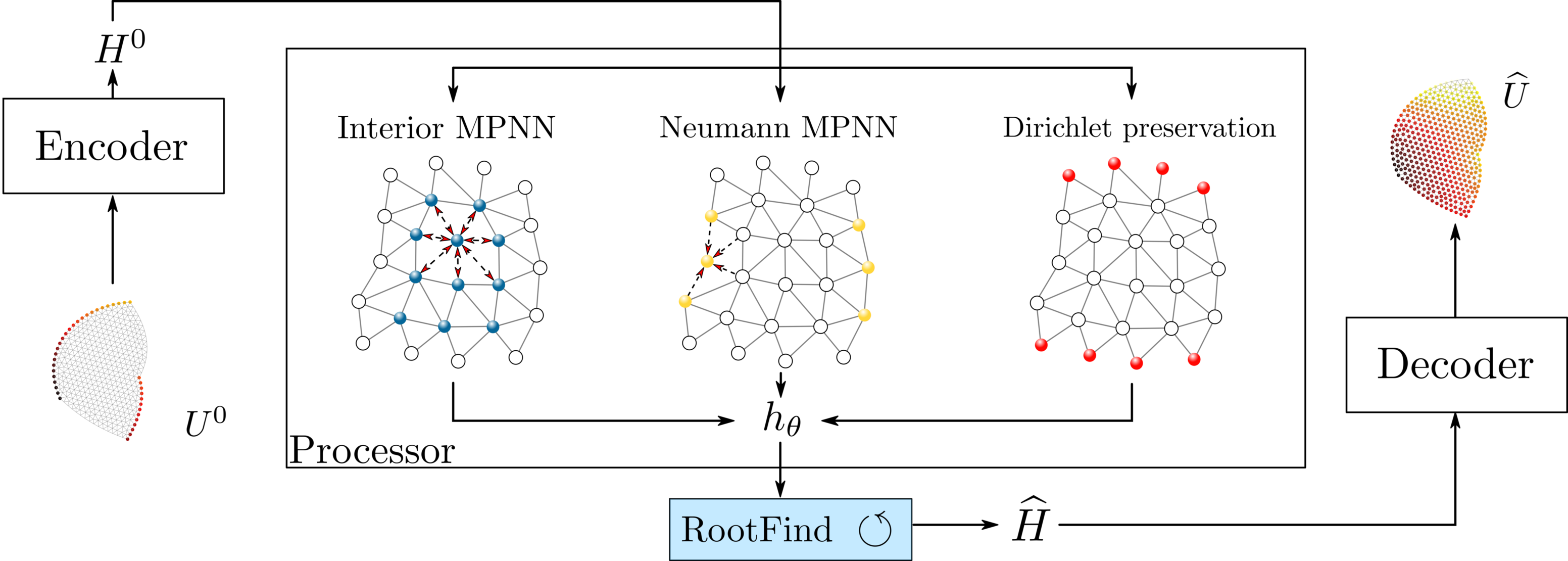}}
\caption{Diagram of \name{}: The model uses an \textit{Encode-Process-Decode} architecture. The encoder maps an initial solution $U^0$ to some latent representation $H^0$. The processor outputs a final latent state $\widehat{H}$ by considering a different treatment for each node type. Dirichlet nodes are preserved during the process, and specific MPNN (red arrows) for Interior and Neumann nodes are computed to build a GNN function $h_\theta$. A black-box ``root-finding" solver automatically propagates the information through the graph by finding the fixed point $\widehat{H}$ of $h_\theta$, starting from the initial guess $H^0$. The decoder maps $\widehat{H}$ back to the physical space to get the final solution $\widehat{U}$.}
\label{sketch_architecture}
\end{center}
\vskip -0.2in
\end{figure*}

Figure \ref{sketch_architecture} gives a global view of \name{}\footnote{code will be available after acceptance}, a Graph Neural Network model with three main components: an \textit{Encoder}, a \textit{Processor}, and a \textit{Decoder}. The ``Encoder-Decoder" mechanism facilitates the connection between the physical space, where the solution lives, and the latent space, where the GNN layers are applied. The Processor is the core component of the model, responsible for propagating the information correctly within the graph. It is specifically designed with two key features: i) it automatically controls the number of Message Passing steps required for convergence; ii) it properly takes into account the boundary conditions by design. \\ 

\noindent \textbf{Encoder} \quad The encoder $E_\theta$, designed as a multilayer perceptron (MLP), maps an initial solution $U^0 \in \mathcal{U}$ to a $d-$dimensional latent state $H^0 \in \mathcal{H}$, $d > 1$. The provided initial solution must fulfil the Dirichlet boundary conditions. This trainable function projects the physical space $\mathcal{U}$ to a higher dimensional latent space $\mathcal{H}$ on which the  GNN layers will be applied.
\\\\
\textbf{Processor} \quad The processor uses a specialized approach for each node type to ensure consistency with the boundary conditions. To propagate the information, the processor constructs a GNN-based function $h_\theta$ that updates both the \textit{Interior} and \textit{Neumann} nodes, effectively capturing the distinct stencils of the discretized Laplace operator. For \textit{Dirichlet} boundary nodes, the corresponding latent variable is kept constant, equal to the imposed value. 
\\\\
\textit{Interior nodes messages} \quad Two separate messages are computed for each node, corresponding to the outgoing and incoming links, using trainable MLPs $\Phi_{\rightarrow,\theta}^\texttt{I}$ and $\Phi_{\leftarrow,\theta}^\texttt{I}$ such that:
\begin{align}
\phi_{\rightarrow,i}^\texttt{I} & = \displaystyle \sum_{j \in \mathcal{N}(i)} \Phi_{\rightarrow,\theta}^\texttt{I}\left(H_i, H_j, d_{ij}, \lVert d_{ij} \rVert \right) 
\label{mp_int_out}  \\
\phi_{\leftarrow,i}^\texttt{I} & = \displaystyle \sum_{j \in \mathcal{N}(i)} \Phi_{\leftarrow,\theta}^\texttt{I} \left(H_i, H_j, d_{ji}, \lVert d_{ji} \rVert \right)
\label{mp_int_in}
\end{align}
where $j \in \mathcal{N}(i)$ stands for all the nodes $j$ in the one-hop neighbourhood of $i$, and $d_{ij}$ and $\lVert d_{ij} \rVert$ represent the relative position vector and the euclidean distance\footnote{For two nodes $i$ and $j$ with coordinates $(x_i,y_i)$ and $(x_j, y_j)$, the relative position vector is $d_{ij} = (x_i - x_j, y_i - y_j)$ and the euclidean distance $\lVert d_{ij} \rVert = \sqrt{(x_i - x_j)^2 + (y_i - y_j)^2}$}.The updated Interior latent variable $\mathbf{z}^\texttt{I} \coloneqq (z_{i}^\texttt{I})_{i\in[N]}$ is computed in a Res-Net fashion of the form:
\begin{align}
z_{i}^\texttt{I} &= H_i + \Lambda_{i,\theta} \left (H_i, b_i, \phi_{\rightarrow,i}^\texttt{I}, \phi_{\leftarrow,i}^\texttt{I} \right)
\label{resnetfashion}
\end{align}
where $\Lambda_{i,\theta}$ is a trainable function designed to stabilize the updating process. The construction of $\Lambda_{i,\theta}$ is detailed in \ref{supplementary_materials_training}. Figure \ref{update_interior} displays the process of updating the Interior node variable. 
\\\\
\textit{Neumann nodes messages} \quad One message from an incoming link is computed and designed to capture the stencil that ensures homogeneous Neumann boundary conditions. It is constructed in a similar manner to (\ref{mp_int_in}), using the MLP $\Phi_{\leftarrow,\theta}^\texttt{N}$ such that:
\begin{align}
\phi_{\leftarrow,i}^\texttt{N} & = \displaystyle \sum_{j \in \mathcal{N}(i)} \Phi_{\leftarrow,\theta}^\texttt{N} \left(H_i, H_j, d_{ji}, \lVert d_{ji} \rVert \right)
\label{mp_neu_in}
\end{align}
The updated Neumann latent variable $\mathbf{z}^\texttt{N} \coloneqq (z_{i}^\texttt{N})_{i\in[N]}$ is computed by combining message (\ref{mp_neu_in}) with problem-related data $b_i$ and the information on the normal vector $n_i$, and passing the result through an MLP $\Psi_\theta$ as follows: 
\begin{align}
z_{i}^\texttt{N} &= \Psi_\theta\left(H_i, b_i, n_i, \phi_{\leftarrow,i}^\texttt{N}\right)
\label{update_neumann}
\end{align}

\begin{figure*}[t!]
\vskip 0.2in
\begin{center}
\centerline{\includegraphics[width=\textwidth]{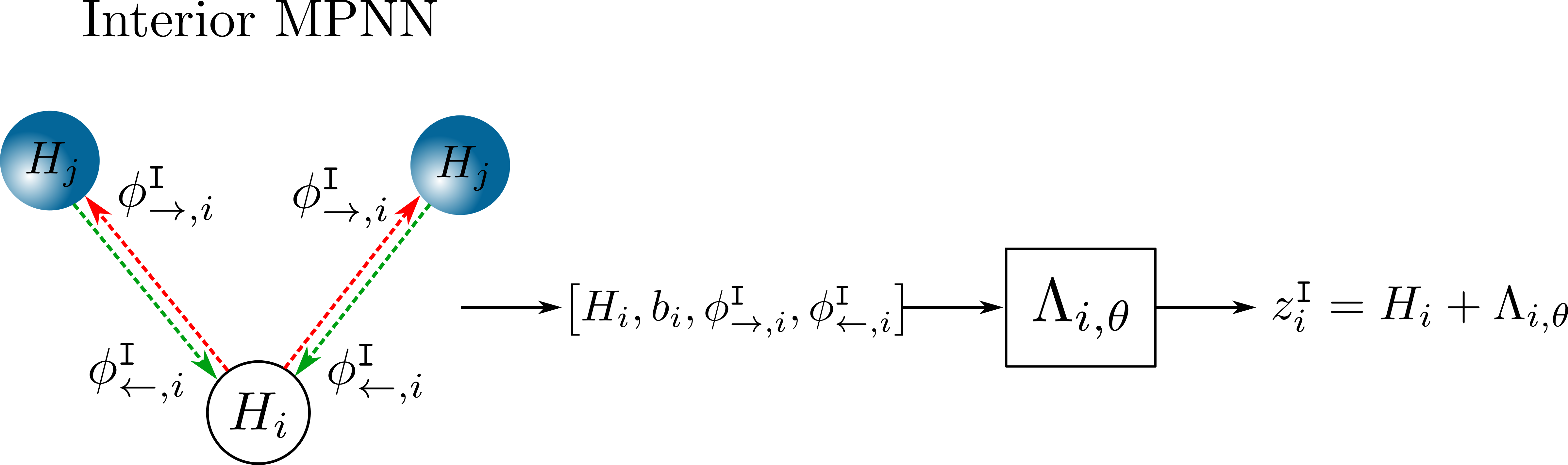}}
\caption{Process of updating the Interior latent variable $z_{i}^\texttt{I}$: Firstly, two MPNNs, $\Phi_{\rightarrow,\theta}^\texttt{I}$ (red arrows) and $\Phi_{\leftarrow,\theta}^\texttt{I}$ (green arrows), are computed to account for the bi-directionality of the edges. These computed messages are then concatenated with problem-specific data $b_i$ and the actual latent state $H_i$ and passed through a trainable function $\Lambda_{i,\theta}$. The output of this trainable function is finally used to calculate $z_{i}^\texttt{I}$ in a Res-Net fashion.}
\label{update_interior}
\end{center}
\vskip -0.2in
\end{figure*}

\textit{GNN-based function} \quad The GNN-based function $h_\theta$, designed to preserve Dirichlet boundary values and separate Interior and Neumann messages, is given by:
\begin{equation}
h_\theta(H,G) = \left \{
\begin{array}{rcl}
H^0 &\qquad& \text{if Dirichlet} \\
\text{LN}\left(\textbf{z}^\texttt{I}\right) &\qquad& \text{if Interior} \\
\text{LN}\left(\textbf{z}^\texttt{N}\right) &\qquad& \text{if Neumann} \\
\end{array}
\right.
\label{eq_func}
\end{equation}
where LN stands for the Layer Normalization operation \citep{layer_normalization}, which consists in normalizing each sample in the minibatch such that the features in the sample have zero mean and unit variance. This operation plays a crucial role in stabilizing $h_\theta$ by constraining its output, resulting in more efficient computation of the subsequent fixed point problem.
\\\\
\textit{Fixed-point problem} \quad One step of message passing only propagates information from one node to its immediate neighbours. In order to propagate information throughout the graph, previous works \citep{ds_gps} repeatedly performed the message passing step, i.e., looped over the function $h_\theta$, either  for a fixed number of iterations or until the problem converges. Iterating until convergence amounts to solving a fixed-point problem
\begin{align}
\widehat{H} = h_\theta\left(\widehat{H},~G\right)
\end{align}
Hence we propose to use a black-box root-finding procedure to directly solve the fixed-point problem:
\begin{align}
\widehat{H} = \text{RootFind}\left(h_\theta\left(H,~G\right) - H \right)
\label{rootfind_eq}
\end{align}
This approach eliminates the need for a predefined number of iterations of $h_{\theta}$ and only requires a threshold precision of the root-finding solver, resulting in a more adaptable and flexible approach. Additional information concerning the motivations behind the use of a fixed point solver can be found in \ref{implicit_models}, particularly in Figure \ref{fig:architecturetypes}. To solve  \eqref{rootfind_eq}, Newton's method is the method of choice, thanks to its fast convergence guarantees. However, to avoid the costly computation of the inverse Jacobian at each Newton iteration, we will use the quasi-Newton Broyden algorithm \citep{broyden}, which uses low-rank updates to maintain an approximation of the Jacobian. It is important to note that, regardless of the chosen root-finding solver, the algorithm starts with the initial guess $H^0$.
\\\\
\textbf{Decoder} \quad The decoder $D_\theta$, designed as an MLP,  maps a final latent variable $\widehat{H} \in \mathcal{H}$ to a meaningful physical solution $\widehat{U} \in \mathcal{U}$. This trainable function is designed as an inverse operation to the encoder. Indeed, the decoder projects back the latent space $\mathcal{H}$ into the physical space $\mathcal{U}$ so that we can calculate the various losses used to train the model.

\subsection{Stabilization}
\label{stabilization}
In Section \ref{architecture}, we modelled a GNN-based network with an ``infinite" depth by using a black-box solver to find the fixed point of function $h_\theta$, enabling for unrestricted information flow throughout the entire graph. 
However, such implicit models suffer from two significant downsides: they tend to be unstable during the training phase and are very sensitive to architectural choices, where minimal changes to $h_\theta$ can lead to large convergence instabilities. The stability of the model around the fixed point $\widehat{H}$ is determined by the  spectral radius $\rho$ of the Jacobian $J_{h_\theta}(\widehat{H})$. Following \citet{deq_stab}, who add a constraint on $\rho$, we add a penalization term (\ref{loss_frob}) in the loss function described in \ref{training}. However, since computing the spectral radius is far too computationally costly, and because the Frobenius norm of the Jacobian is an upper bound for its spectral radius, we adopt the method outlined in \citet{deq_stab}, which estimates this Frobenius norm using the Hutchinson estimator \citep{hutchinson}. By doing so, we build a model that satisfies, post-training, $\rho < 1$: the function $h_\theta$ becomes contractive, thus ensuring the global asymptotic convergence of the model. As a result, during inference, one can simply iterate on the Processor (in an RNN-like manner) until convergence (i.e. loop over the $h_\theta$ function). More importantly, the model could work with any kind of root-finding solver. Overall, this approach offers strong convergence guarantees and addresses the stability issues commonly encountered in implicit models.

\subsection{Training}
\label{training}

\textbf{Loss function} \quad 
The entire \name{} model is trained by minimizing the following cost function: 
\begin{align}
    \mathcal{L} ~=~ & \mathcal{L}_\text{res}(\widehat{U}, G) \label{loss_base} \\
    & + \lambda \times MSE\left(\widehat{U} - U^\text{ex}\right) \label{loss_sup} \\
    & + \beta \times ||J_{h_\theta}(\widehat{H})||_\text{F} \label{loss_frob} \\
    & + \text{MSE}\left(E_\theta(\widehat{U}) - \widehat{H}\right) \label{loss_enc} \\
    & + \text{MSE}\left(D_\theta(E(\widehat{U})) - \widehat{U}\right) \label{loss_dec} 
\end{align}
Line (\ref{loss_base}) represents the residual loss described in (\ref{loss-function}), and line (\ref{loss_sup}) is an additional supervised loss ($U^{\text{ex}}$ being the LU ground truth and $\lambda$ a small weight, see Section \ref{results}). Line (\ref{loss_frob}) is the regularizing term defined in \ref{stabilization}. Lines (\ref{loss_enc}) and (\ref{loss_dec}) are designed to learn the autoencoding mechanism together: Line (\ref{loss_enc}) aims to properly encode a solution while Line (\ref{loss_dec}) steers the decoder to be the inverse of the encoder. To handle the minimization of the overall structure, a single optimizer is used with two different learning rates, one for the autoencoding process and one for the Message Passing process. This ensures that the autoencoding process is solely used for the purpose of bridging the physical and latent spaces, with no direct impact on the accuracy of the computed solution.
\\\\
\textbf{Backpropagation} \quad The training of the proposed model has been found to be computationally intensive or even not feasible when backpropagating through all the operations of the fixed-point solver. However, using the approach outlined in Theorem 1 in \citet{deq} significantly enhances the training process by differentiating directly at the fixed point, thanks to the implicit function theorem. This methodology requires the resolution of two fixed-point problems, one during the inference phase and the other during the backpropagation phase. In contrast to traditional methods, this approach removes the need to open the black-box, and only requires constant memory. Supplementary materials concerning the regularization, the training process and the architecture and implementation of \name{} can be found in \ref{supplementary_materials_training} and \ref{implicit_models}.

\section{Theoretical properties}
\label{theory}

This section investigates several theoretical properties of the proposed approach. Specifically, we demonstrate that \name{} satisfies a property of universal approximation, i.e. the model is able to approximate the optimal solution to the considered statistical problem \eqref{ssp-problem} up to any arbitrary precision, provided the network is large enough. Following the approach and notations outlined in Section \ref{statistical_problem}, a discretized Poisson problem $E_h = \left( \Omega_h,~A,~B\right)$ can be seen as a graph problem $G = \left(N,~A,~B\right)$, where $A$ and $B$ are respectively the interaction and individual terms on the $N$ nodes of the graph $G$. The original problem at hand searches for an optimal solution $U^\star_G \in \mathbb{R}^N$ as follows:

\begin{equation}
\displaystyle U^\star_G = \underset{U \in \mathbb{R}^N}{\text{argmin}}~\mathcal{L}_\text{res}\left(U,~G\right)
\label{eq:psignn_optiproblem}
\end{equation}

However, instead of searching directly for $U^\star_G$, we seek for a function $h^\star_G :\mathbb{R}^N \rightarrow ~ \mathbb{R}^N$ whose fixed point is $U^\star_G$:

\begin{equation}
\displaystyle h^\star_G = \underset{h:~\mathbb{R}^N \rightarrow ~ \mathbb{R}^N}{\text{argmin}}~ \mathcal{L}_\text{res}\left(\text{FixedPoint}(h),~G\right)
\label{eq:psignn_optiproblemfixedpoint}
\end{equation}

The first step in proving the consistency of the approach is to determine whether problems \eqref{eq:psignn_optiproblem} and \eqref{eq:psignn_optiproblemfixedpoint} are equivalent, i.e., if solving \eqref{eq:psignn_optiproblemfixedpoint} yields the solution to the original problem \eqref{eq:psignn_optiproblem}.

\begin{prop}[Equivalence of direct and fixed-point formulations]
\label{prop:equivalence}\mbox{}\\
Problems \eqref{eq:psignn_optiproblem} and \eqref{eq:psignn_optiproblemfixedpoint} are equivalent, i.e., for any problem $G$, any solution $U^\star_G$ of \eqref{eq:psignn_optiproblem} can be turned into a solution $h^\star_G$ of \eqref{eq:psignn_optiproblemfixedpoint} and vice versa.
\end{prop}

\begin{proof}
See \ref{pf:equivalence} in \ref{appendix:theory}.
\end{proof}

The next step investigates whether the \name{} architecture is able to find an approximation of $h^\star_G$. To address this, we begin by proving that problem \eqref{eq:psignn_optiproblem} satisfies the hypotheses of Corollary $1$ of DSS - \citet{deep_statistical_solver}.

\begin{prop}[Satisfying the hypotheses of Corollary 1 in DSS]
\label{prop:satisfyconditions}\mbox{}\\
Problem \eqref{eq:psignn_optiproblem}, which can be rewritten as searching for the function:

\begin{equation*}
\varphi: \;\;\; 
\begin{array}{rcl}
\mathcal{S} & \rightarrow & \mathbb{R}^N \\ G = (N,~A,~B) &\;\mapsto\; &  U^\star(G):=  \underset{U}{\argmin}~\mathcal{L}_\text{res}(U,~G)
\end{array}
\end{equation*}

satisfies the hypothesis of Corollary $1$ in \citet{deep_statistical_solver}.  
\end{prop}

\begin{proof}
See \ref{pf:satisfyconditions} in \ref{appendix:theory}.
\end{proof}

Let $\mathcal{H}$ be the space of functions that can be realized by Graph Neural Networks, as defined in \citet{deep_statistical_solver}. The previously mentioned Corollary $1$ in \citet{deep_statistical_solver} establishes the existence of a network with a GNN-based architecture, denoted as $\widehat{\varphi} \in \mathcal{H}$, that can approximate $\varphi: G \mapsto U^\star_G$ with arbitrary precision for any given graph $G$ from the considered problem distribution.  We have the following corollary:

\begin{cor}[Existence of a GNN model approximating $\varphi$]
\label{cor:exsitence_gnn}\mbox{}\\
For any $\varepsilon > 0$, there exists a GNN-based model $\widehat\varphi \in \mathcal{H}$ such that for any problem $G$ from our problem distribution $\mathcal{D}$:
\begin{equation*}
  \| \widehat\varphi(G) - \varphi(G)\| \leqslant \varepsilon   
\end{equation*}
\end{cor}

Based on Proposition \ref{prop:equivalence} and its proof, this result can be extended to approximate the solution of \eqref{eq:psignn_optiproblemfixedpoint}, yielding the following universal approximation property for our \name{} method:

\begin{thm}[Universal Approximation Property]
\label{thm:universal_approx}\mbox{}\\
For any precision $\varepsilon > 0$, there exists a parameterization $\theta_\varepsilon$ of a \name{} architecture with sufficiently large layers, namely, a function $\text{\name{}}_{\theta_\varepsilon}: \mathcal{S} \rightarrow \mathbb{R}^N$, which, for any problem $G$ (i.e., for any mesh, boundary conditions and force terms), approximates the optimal solution of problem \eqref{eq:psignn_optiproblem} with precision less than $\varepsilon$: 

\begin{equation*}
    \forall ~G, ~~~ \left\| \text{\name{}}_{\theta_\varepsilon}(G) - \varphi(G) \right\|  \leqslant \epsilon    
\end{equation*}

\end{thm}

\begin{proof}
See \ref{pf:theorem} in \ref{appendix:theory}.
\end{proof}

Lastly, the set of functions $h$ that admit a unique fixed point close to $\widehat\varphi_{\text{GNN}}(G)$ (see \ref{pf:theorem}) can be big (although all these solutions lead approximately to the same fixed point). It is possible to select an optimal solution within this space that is also contractive. This can be achieved in a Lagrangian spirit instead of a supplementary constraint by introducing a penalty term to the loss function, as shown in \eqref{loss_frob}. However, it is important to note that the resulting function can only be assumed to be predominantly contractive with respect to $H$:

\begin{prop}[Contractivity of $h_{\theta\varepsilon}$]
\label{prop:contractivity}\mbox{}\\
For any precision $\varepsilon > 0$, the function $h_{\theta_\varepsilon}(H,G)$ in the Processor of the \name{} architecture obtained by Theorem \ref{thm:universal_approx} can be assumed to be contractive with respect to $H$ for any pair of points farther than $\sqrt{\varepsilon}$.\\
\end{prop}

\begin{proof}
See \ref{pf:contractivity} in \ref{appendix:theory}.  
\end{proof}

\section{Experiments, Results, and Discussion}
\label{results}

This section presents an in-depth evaluation of \name{} based on experiments on synthetic data. The first Section \ref{results:common_setup} presents an overview of the synthetic dataset used in these experiments and introduces the standard experimental setup. The performance evaluation of the proposed approach is then performed in three other sections: \ref{results:direct_comparison} is focused on addressing Poisson problems with Dirichlet boundary conditions only, enabling a direct comparison with the state-of-the-art Deep Statistical Solvers (DSS) method, while \ref{results:mixed_bc} is focused on Poisson problems with mixed boundary conditions. Furthermore, Section \ref{subsec:additionalresults} proposes various generalization tests to conduct a thorough assessment of \name{} performance, showcasing its originality. Finally, Section \ref{subsec:complexity} investigates the inference complexity of the proposed model.

\subsection{Dataset \& Experimental setup}
\label{results:common_setup}

\begin{figure*}[t]
\vskip 0.2in
\begin{center}
\centerline{\includegraphics[width=\textwidth]{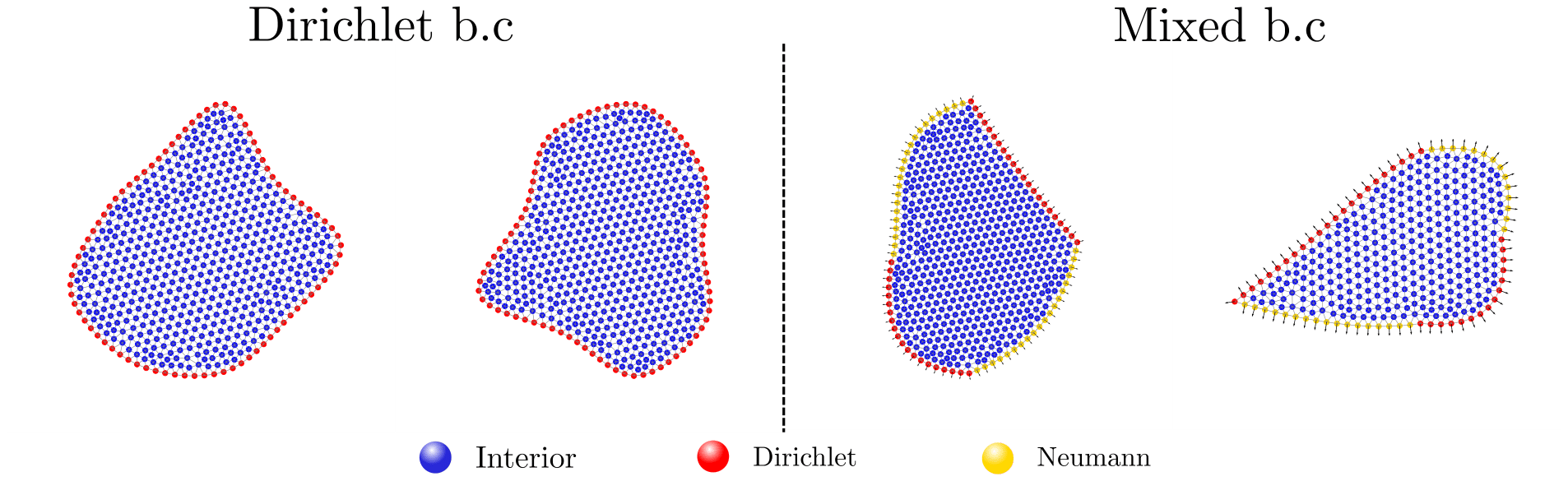}}
\caption{Geometries extracted from the synthetic dataset colored with respect to their node type (blue: Interior, red: Dirichlet, yellow: Neumann). These geometries are used to solve Poisson problems with Dirichlet boundary conditions only (on the left side) and mixed boundary conditions (on the right side). The arrows on the boundary of the two domains on the right represent the normal vectors.}
\label{fig:dataset}
\end{center}
\vskip -0.2in
\end{figure*}

\begin{figure}[t]
     \centering
     \begin{subfigure}[b]{0.45\textwidth}
         \centering
         \includegraphics[scale=0.45]{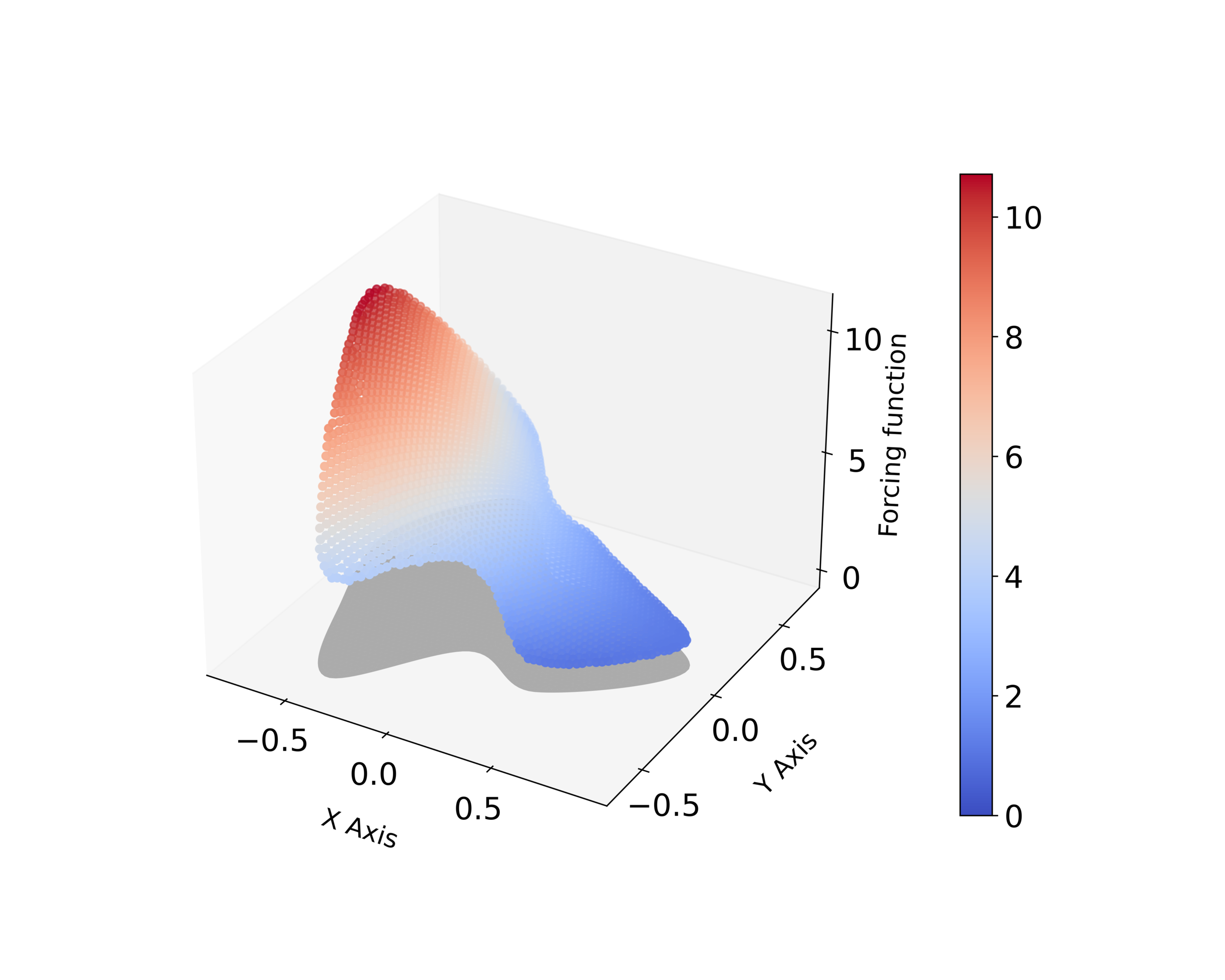}
         \caption{Forcing function}
         \label{fig:forcingfunction}
     \end{subfigure}
     \hfill
     \begin{subfigure}[b]{0.5\textwidth}
         \centering
         \includegraphics[scale=0.45]{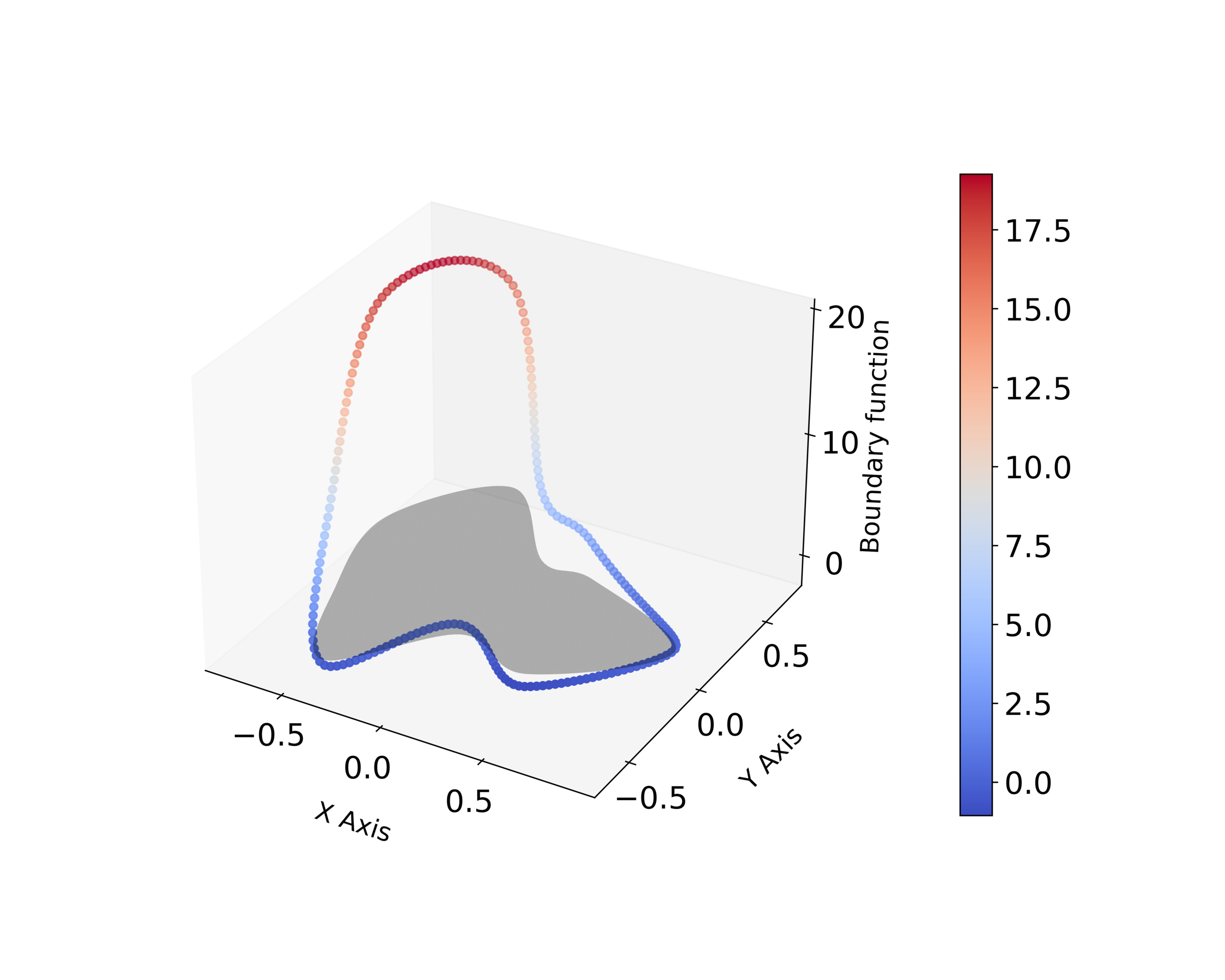}
         \caption{Boundary function}
         \label{fig:boundaryfunction}
     \end{subfigure}
     \caption{Figures \ref{fig:forcingfunction} and \ref{fig:boundaryfunction} display the discrete values of a forcing term $f$ and a boundary function $g$, on a mesh sampled from the test set. The coefficients of $f$ and $g$, uniformly sampled in $[-10, 10]$, are $r_1 = 3.2$, $r_2 = -7.5$, $r_3 = 1.1$, $r_4 = 5.7$, $r_5 = -9.5$, $r_6 = 0.47$, $r_7 = -8.8$, $r_8 = 9.11$, and $r_9 = 3.5$. The considered mesh is shown as a grey shadow in the plot.}
     \label{fig:plotfunction}
\end{figure}

\textbf{Synthetic dataset} \quad The dataset used in all experiments consists of  $6000$ / $2000$ / $2000$ training/validation/test samples of  Poisson problems \eqref{poisson-equation} generated as follows. Random $2$D domains $\Omega$ are generated using $10$ points, randomly sampled in the unit sphere. These points are then connected using Bezier curves to form the boundary of the domain $\Omega_h$. The ``MeshAdapt" mesher from GMSH \citep{gmsh} is used to discretize $\Omega$ into an unstructured triangular mesh $\Omega_h$. Our analysis includes several datasets to assess the performance of the model, which differ regarding the type of boundary conditions. For each dataset, the meshes have approximately $500$ nodes (the automatic mesh generator does not include precise control of the number of nodes $N$). When considering Poisson problems with Dirichlet boundary conditions only,  Dirichlet boundary nodes are applied to the entire boundary of the meshes in the dataset (left part in Figure \ref{fig:dataset}). When considering Poisson problems with mixed boundary conditions (\ref{poisson-equation}), the boundary is randomly divided into four sections and Dirichlet boundary conditions are applied on two opposite sections, while Neumann boundary conditions are imposed on the remaining opposite sections (right part of Figure \ref{fig:dataset}). Forcing functions $f$ and boundary functions $g$ from (\ref{poisson-equation}) are defined as random quadratic polynomials with coefficients sampled from uniform distributions such as:
\begin{align}
f(x,y) & = r_1(x-1)^2 + r_2 y^2 + r_3 & (x,y) & \in \Omega \\
g(x,y) & = r_4x^2 + r_5y^2 + r_6xy + r_7x + r_8y + r_9 & (x,y) &\in \partial \Omega 
\end{align}
where $r_{i \in \left[1, \cdots, 9\right]}$ are randomly sampled in $[-10,10]$. Figure \ref{fig:plotfunction} illustrates examples of such $f$ and $g$ functions on a sampled mesh. \\

\textbf{Metrics} \quad Throughout these experiments, we consider the solution of the discretized Poisson problem given by the classical LU decomposition method as the ``ground truth". The reported metrics are the residual loss \eqref{loss-function} and the mean squared error (MSE) between the output of the model and the LU solution.
\\\\
\textbf{Common setup} \quad \name{} is implemented in Pytorch using the Pytorch-Geometric library \citep{pytorch_geometric} to handle graph data. The dimension $d$ of the latent space $\mathcal{H}$ is set to $10$. Each neural network block in the architecture (equations \ref{mp_int_out} to \ref{update_neumann}) has one hidden layer of dimension $10$ with a ReLU activation function. All model parameters are initialized using Xavier initialization \citep{xavier_init}. For the training, the provided initial solution is set to zero everywhere except at the Dirichlet nodes, which are assigned to the corresponding discrete value of $g$. Gradient clipping is employed to prevent exploding gradient issues and set to $10^{-2}$. The model requires, at each iteration, the solution of two fixed point problems, one for the forward pass and one for the backward pass, as outlined in Section \ref{training}. These problems are solved using Broyden's method with a relative error as the stopping criteria. The latter is set to $10^{-5}$ with a maximum of $500$ iterations for the forward pass and to $10^{-8}$ with a maximum of $500$ iterations for the backward pass.

\subsection{Poisson problems with Dirichlet boundary conditions}
\label{results:direct_comparison}

This section aims to assess the performance of \name{} for solving Poisson problems with Dirichlet boundary conditions, allowing for direct comparison with the state-of-the-art Deep Statistical Solvers (DSS) model.
\\\\
\textbf{Deep Statistical Solvers} \quad As outlined in Section \ref{related_work}, our study proposes to evaluate the efficiency of \name{} by conducting a comparative analysis with the state-of-the-art Deep Statistical Solvers (DSS) model \citep{deep_statistical_solver}. The authors of the DSS model have demonstrated that the convergence of their model is reliant upon a number of MPNNs that is directly proportional to the diameter\footnote{The shortest path (node-wise) between the two farthest nodes in a mesh.} of the meshes considered within the dataset. The architecture of the DSS model can be described as follows: Initially, a latent state initialized to $0$ everywhere is assigned to each node within the network. This latent state then undergoes a fixed sequence of MPNNs, illustrated schematically in Figure \ref{fig:archidss}. At each layer, a multilayer perceptron (MLP) is employed to decode the latent state, transforming it into a physical state, upon which an intermediate residual loss (\ref{loss-function}) is calculated. The final training loss is computed as a weighted cumulative sum of all intermediate losses. It is noteworthy that the authors of \citet{deep_statistical_solver} have compared their model trained with either a residual \eqref{loss-function}
or an MSE loss, demonstrating enhanced generalization capabilities when using a residual loss function. Additionally, we recall several drawbacks of the DSS model: i) the fixed number of MPNNs hampers its ability to generalize across various mesh sizes, ii) the weights associated with the MPNNs and the MLP decoder differ at each layer, resulting in a growing model if the number of iterations needed for convergence increases, iii) it exclusively addresses Poisson problems with Dirichlet boundary conditions, and iv) no explicit treatment of boundary conditions is considered in the model architecture. With the introduction of \name{}, we propose solutions to address these challenges.
\\\\
\textbf{Experimental setup} \quad In this experiment, both \name{} and DSS are trained and tested using a dataset of Poisson problems with Dirichlet boundary conditions only, generated following the process described in Section \ref{results:common_setup}. Note that this dataset is similar to the one used in the original DSS study, allowing for fair comparison. \name{} follows the architectural specifications outlined in Section \ref{architecture}, with the exception of excluding functions related to Neumann nodes. During training, the model is optimized using the loss function \ref{loss_base}, with $\lambda = 0$ (no additional "ground-truth" MSE minimization) and $\beta = 1$. Training is done using Nvidia P$100$ GPUs and the Adam optimizer with its default Pytorch hyperparameters, except for the initial learning rate, which was set to $0.05$ for the autoencoding process and $0.01$ for the main process, as discussed in Section \ref{training}. The {\em ReduceLROnPlateau} scheduler from Pytorch is used to progressively reduce the learning rate from a factor of $0.5$ during the process, enhancing the training. The remaining hyperparameters adhere to the common setup detailed in Section \ref{results:common_setup}. Regarding the DSS model, it remains consistent with the original study, i.e. trained with $30$ layers. \\

\begin{figure}[t!]
     \centering
     \begin{subfigure}[b]{0.3\textwidth}
         \centering
         \includegraphics[width=\textwidth]{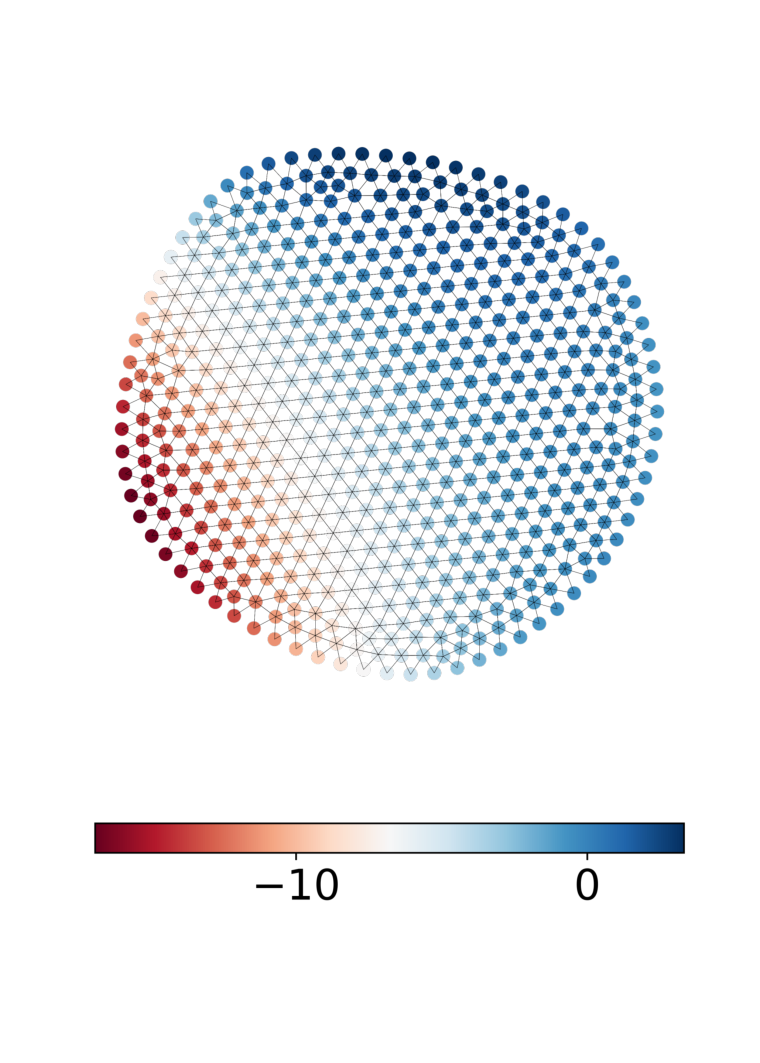}
         \caption{Data-driven solution}
         \label{subfig:psignn_data_driven_solution}
     \end{subfigure}
     \hfill
     \begin{subfigure}[b]{0.3\textwidth}
         \centering
         \includegraphics[width=\textwidth]{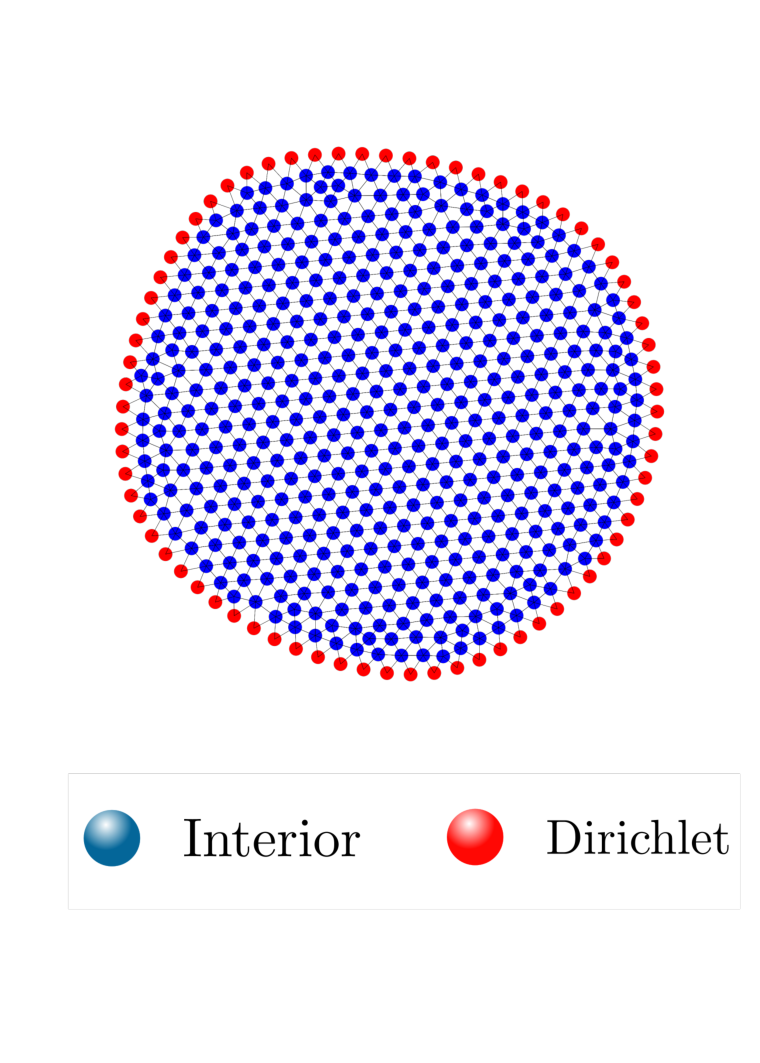}
         \caption{Node types}
         \label{subfig:psignn_node_types}
     \end{subfigure}
     \hfill
     \begin{subfigure}[b]{0.3\textwidth}
         \centering
         \includegraphics[width=\textwidth]{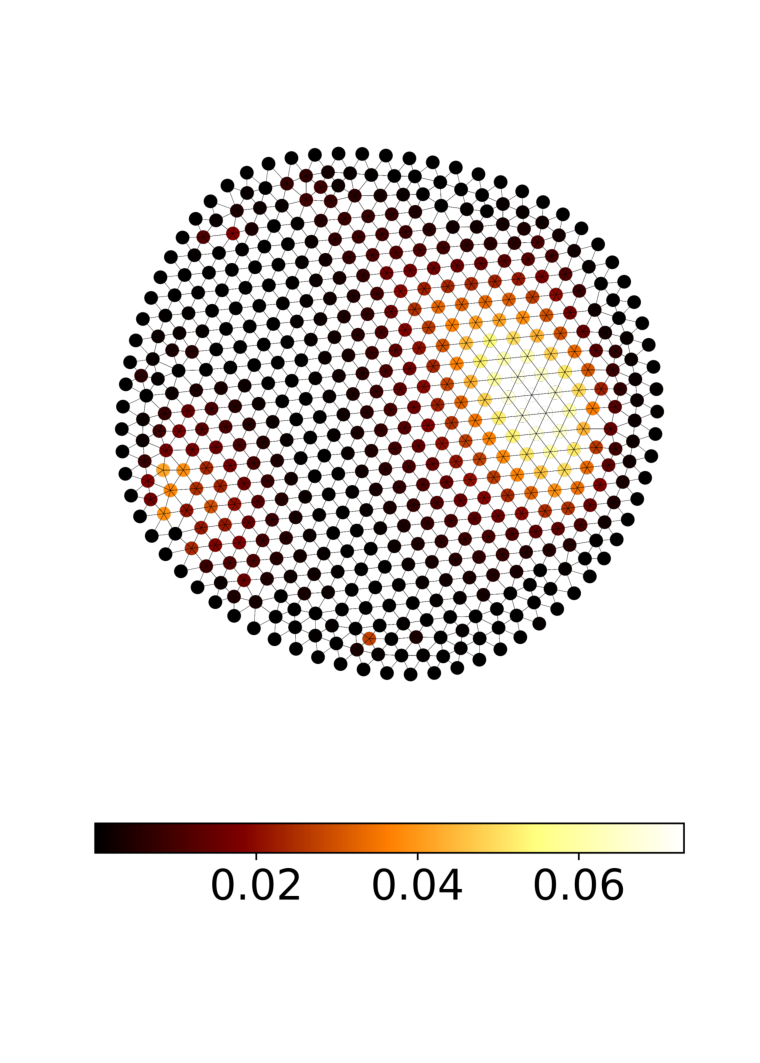}
         \caption{Error map}
         \label{subfig:psignn_error_map}
     \end{subfigure}
     \hfill \par \bigskip
     \begin{subfigure}[b]{\textwidth}
         \centering
         \includegraphics[width=\textwidth]{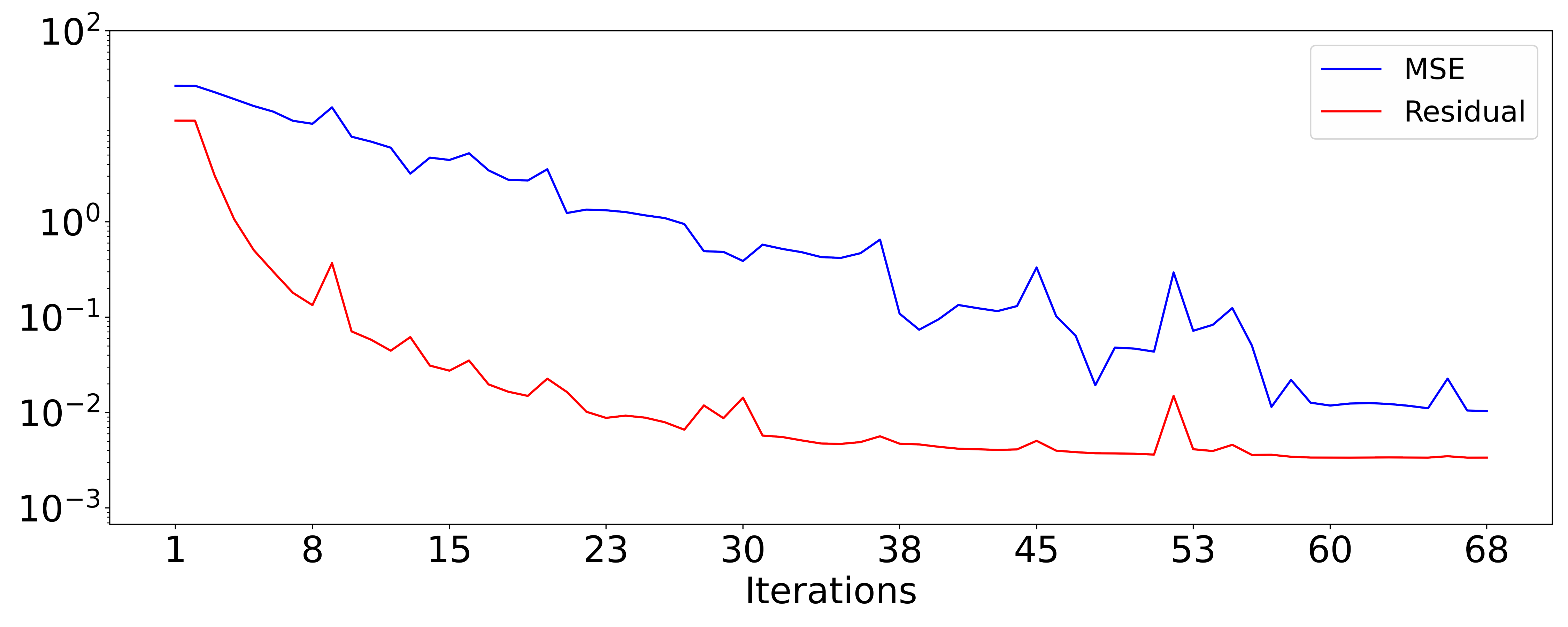}
         \caption{Residual (red) and MSE w/LU (blue) losses}
         \label{subfig:psignn_losses_evol}
     \end{subfigure}
     \caption{Illustration of the resolution of a Poisson problem extracted from the test set using \name{}. Figure \ref{subfig:psignn_data_driven_solution} shows the data-driven solution obtained at the last iteration, while Figure \ref{subfig:psignn_error_map} displays the map of squared errors between the data-driven solution and the LU solution. Figure \ref{subfig:psignn_node_types} illustrates the different types of nodes. At the bottom, Figure \ref{subfig:psignn_losses_evol} depicts the evolution of the Residual (in red) and MSE w/LU (in blue) across the $68$ iterations of the model.}
     \label{fig:dirichletsample}
\end{figure}

\textbf{Results Analysis} \\\\
Table \ref{tab:dirichletVDS} presents the residual and MSE w/LU averaged over the entire test set for \name{} and DSS. From this table, it is possible to observe that DSS provides slightly better results than \name{} in terms of residual. This is because, in \name{}, the training loss not only minimizes the residual but also integrates several other components within the training procedure, such as the stabilization process, which imposes additional constraints on the model weights. Additionally, in DSS, the training loss is computed as a weighted sum of intermediate residual losses, calculated at each iteration of the model, while in \name{}, the residual loss is computed only at the last layer due to the implicit nature of the iterations. However, the results take a different flavor when looking at the MSE w/LU. There, we can see that \name{} outperforms DSS. This can be explained by the ability of the model to automatically adjust its number of Message-Passing steps, as well as the autoencoding mechanism that better encodes and decodes Dirichlet boundary conditions up to a precision of order $10^{-6}$. This enhanced feature allows for better information flow since the initial latent state is properly initialized and results in a better MSE w/LU. Additionally, \name{} is able to outperform DSS with only $1444$ parameters, in contrast to the DSS model, which requires $36930$ iterations (performing only $30$ iterations!). Finally, \name{} is trained on the same dataset as DSS, which contains meshes of approximately $500$ nodes. However, \name{}, even though it is trained on fixed-size meshes, has the remarkable capability of adjusting its iteration count by itself to handle meshes of varying sizes, a feature analyzed in detail in Section \ref{subsec:additionalresults}. \\\\
Figure \ref{fig:dirichletsample} illustrates the resolution with \name{} of a test instance, a Poisson problem with $527$ nodes. The figure presents the evolution of the Residual (in the loss function) and the MSE w/LU (though it was not used during training) along the $68$ iterations of the Broyden algorithm. At convergence, the Residual reaches a value of $3.36$e-$3$ and an MSE w/LU of $1.03$e-$2$. Similar results on all test examples validate the \name{} approach, showcasing its ability to solve Poisson problems with Dirichlet boundary conditions. 

\begin{table}[t!]
    \renewcommand{\arraystretch}{1.5}
    \centering
        \begin{tabular}{l@{\hspace{1.1cm}}c@{\hspace{1.1cm}}c@{\hspace{1.1cm}}c}
            \hline
            & Residual ($10^{-3}$) & MSE w/LU ($10^{-2}$) & Nb of weights\\
            \hline 
            \name{} & 2.69 $\pm$ 0.4 & \textbf{0.85 $\pm$ 2} & \textbf{1444} \\
            \hline 
             DSS & \textbf{0.23 $\pm$ 0.2} & 3.0 $\pm$ 2 & 36930 \\
            \hline 
            \end{tabular}
    \caption{Results of \name{} and DSS averaged over the whole test set}
    \label{tab:dirichletVDS}
\end{table}

\subsection{Poisson problems with mixed boundary conditions}
\label{results:mixed_bc}

This section aims to assess the performance of \name{} when extended to solve Poisson problems with mixed boundary conditions.\\

\textbf{Experimental setup} \quad In this experiment, \name{} is trained and tested using a dataset of Poisson problems with mixed boundary conditions, generated following the process described in Section \ref{results:common_setup}. Similar to the previous test case, training is performed using Nvidia P$100$ GPUs and the Adam optimizer with its default PyTorch hyperparameters. However, the initial learning rates differ: $0.01$ is used for the autoencoding process, while $0.005$ is used for the main process. To enhance the training, we employ the PyTorch {\em ReduceLROnPlateau} scheduler, gradually reducing the learning rate by a factor of $0.8$ during the process. The full \name{} architecture, as described in \ref{architecture}, is trained using the loss function \ref{loss-function} with $\lambda = 0.001$ and $\beta = 1$. The remaining hyperparameters are set according to the common framework described in \ref{results:common_setup}.\\

\textbf{Results Analysis}\\

Table \ref{tab:neumannresults} presents the averaged Residual and MSE w/LU errors obtained by \name{} on the entire test set. These results are comparable in terms of Residual and slightly higher in terms of MSE w/LU compared to those obtained in Section \ref{results:direct_comparison}. This disparity can be attributed to the increased difficulty of learning problems with Neumann boundary conditions, which have higher conditioning. As a result, even with the same residual error, the MSE w/LU can vary due to the challenging conditioning of the mixed boundary condition problem. However, these results demonstrate the accuracy of \name{} in solving Poisson problems with mixed boundary conditions. Notably, the model has $2175$ parameters, which is slightly more than the one developed in Section \ref{results:direct_comparison}, due to the additional networks required to take into account the Neumann boundary conditions.\\

\begin{figure}[t!]
     \centering
     \begin{subfigure}[b]{0.3\textwidth}
         \centering
         \includegraphics[width=\textwidth]{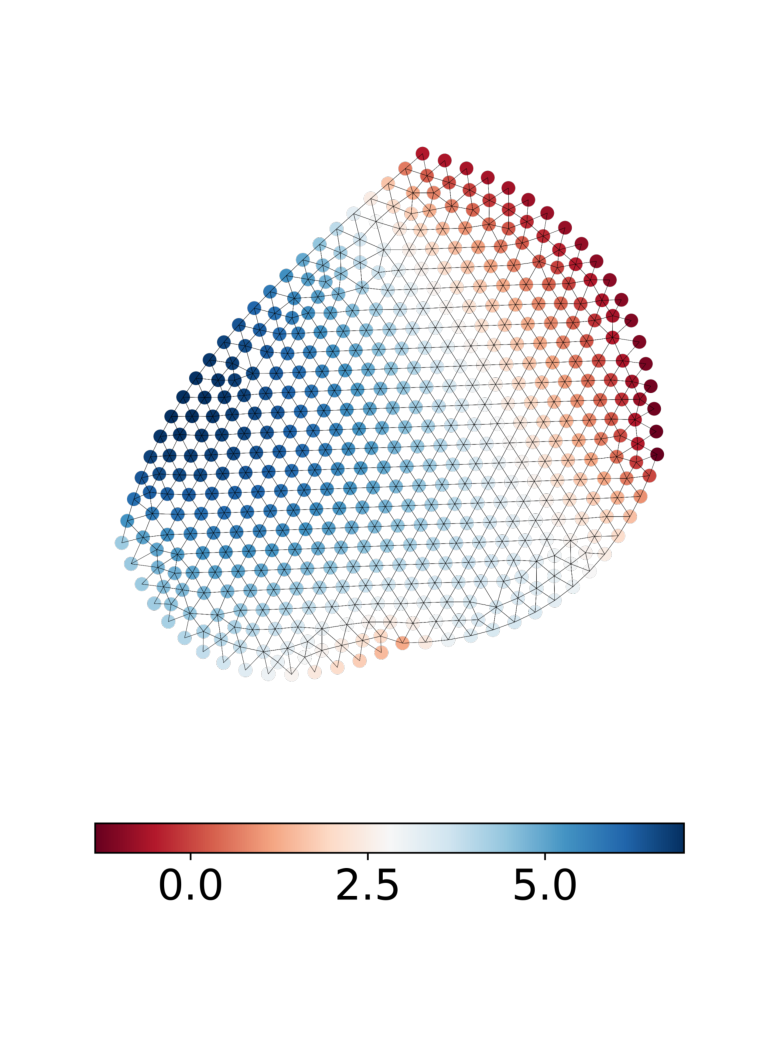}
         \caption{Data-driven solution}
         \label{subfig:psignn_neumann_data_driven_solution}
     \end{subfigure}
     \hfill
     \begin{subfigure}[b]{0.3\textwidth}
         \centering
         \includegraphics[width=\textwidth]{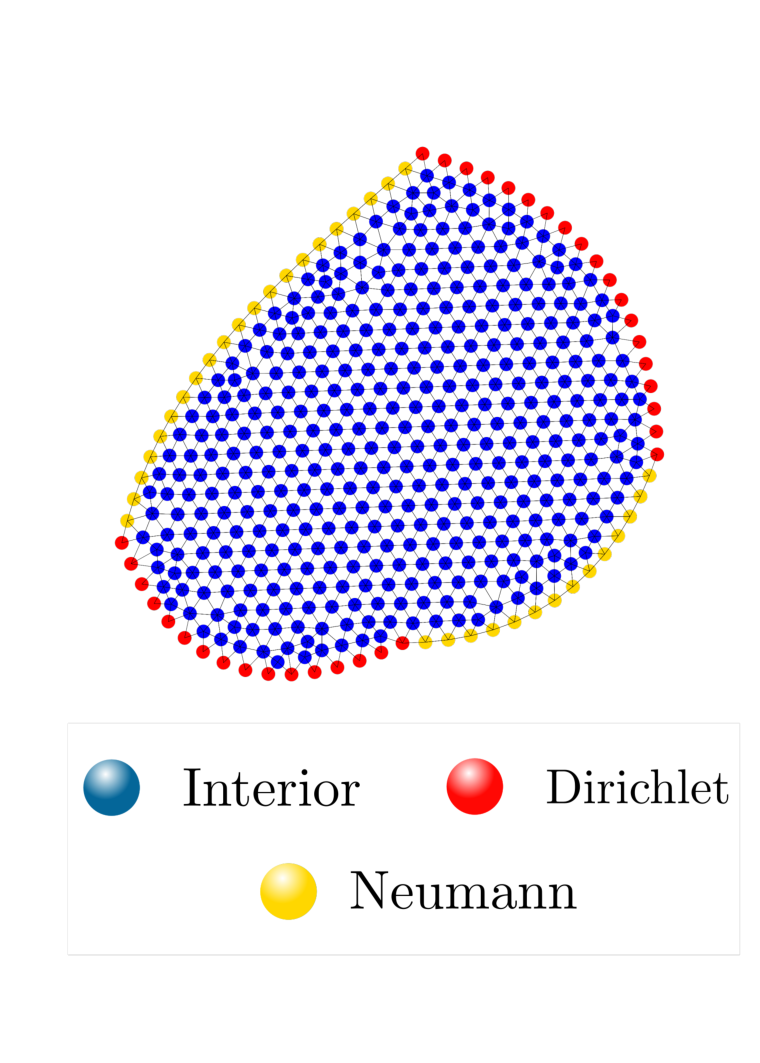}
         \caption{Node types}
         \label{subfig:psignn_neumann_node_types}
     \end{subfigure}
     \hfill
     \begin{subfigure}[b]{0.3\textwidth}
         \centering
         \includegraphics[width=\textwidth]{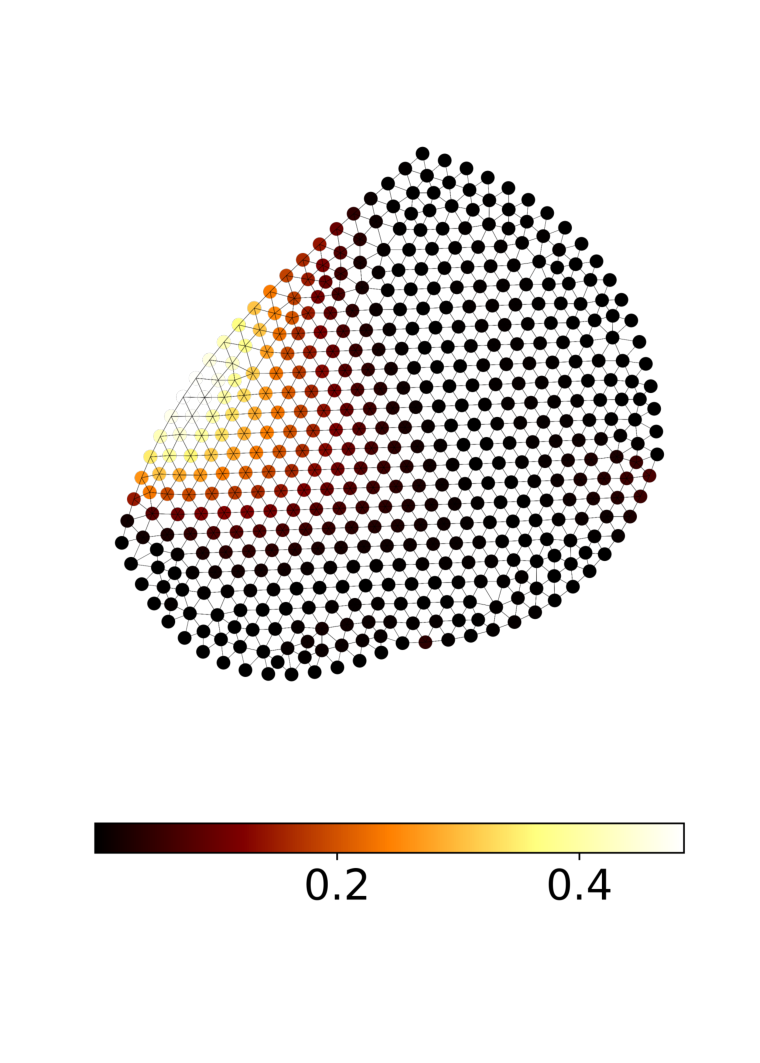}
         \caption{Error map}
         \label{subfig:psignn_neumann_error_map}
     \end{subfigure}
     \hfill \par \bigskip
     \begin{subfigure}[b]{\textwidth}
         \centering
         \includegraphics[width=\textwidth]{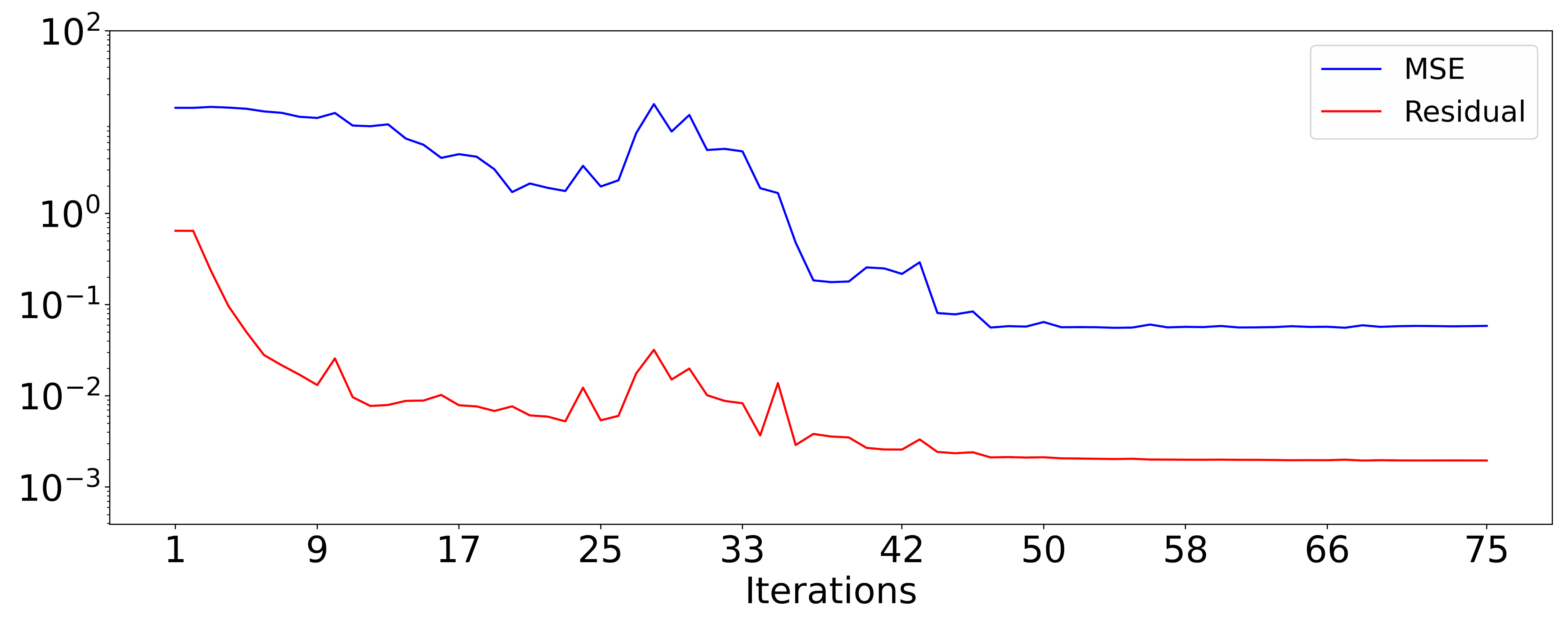}
         \caption{Residual (red) and MSE w/LU (blue) losses}
         \label{subfig:psignn_neumann_losses_evol}
     \end{subfigure}
     \caption{Illustration of the resolution of a Poisson problem extracted from the test set using \name{}. Figure \ref{subfig:psignn_neumann_data_driven_solution} shows the data-driven solution obtained at the last iteration, while Figure \ref{subfig:psignn_neumann_error_map} displays the map of squared errors between the data-driven solution and the LU solution. Figure \ref{subfig:psignn_neumann_node_types} illustrates the different types of nodes. At the bottom, Figure \ref{subfig:psignn_neumann_losses_evol} depicts the evolution of the Residual (in red) and MSE w/LU (in blue) across the $75$ iterations of the model.}
     \label{fig:sampleneumann}
\end{figure}

Figure \ref{fig:sampleneumann} displays the solution to a specific problem with $470$ nodes extracted from the test set. At convergence, the model reaches a Residual of $1.9$e-$3$ and an MSE w/LU of $5.8$e-$2$, showcasing the effectiveness of the method on this test sample. The autoencoding process ensures accurate encoding and decoding of the Dirichlet boundary conditions, thereby preserving them throughout the iterations with an error of magnitude $10^{-6}$. The error map reveals that the highest errors are primarily concentrated near the Neumann boundary nodes, aligned with expectations. As discussed in Section \ref{statistical_problem}, the graphs considered in this study are directed from the Dirichlet boundary nodes toward the interior of the graph. Consequently, the flow of information is propagated from these nodes towards the inner region of the domain. However, in the case of Neumann nodes, which involve bidirectional edges, the task of propagating information across the entire graph becomes more challenging compared to the fully Dirichlet problem. This is because, in the full Dirichlet problem, information can flow from the entire boundary of the domain. However, in the presence of Neumann nodes, the bidirectional edges complicate the propagation of information throughout the graph, and the Neumann values must be gradually approached. This is in contrast to Dirichlet values, which are considered ``exact''. As a result, the model requires more iterations to attain the solution. 

\begin{table}[!t]
    \renewcommand{\arraystretch}{1.5}
    \centering
        \begin{tabular}{l@{\hspace{1.1cm}}c@{\hspace{1.1cm}}c@{\hspace{1.1cm}}c}
            \hline
            Metrics & Residuals ($10^{-3}$) & MSE w/LU & Nb of weights\\
            \hline 
            \name{} & $3.16$ $\pm$ $0.2$ & $0.15$ $\pm$ $0.04$ & $2175$ \\
            \hline
        \end{tabular}
    \caption{Results of \name{} averaged over the whole test set.}
    \label{tab:neumannresults}
\end{table}

\subsection{Sensitivity analyses}
\label{subsec:additionalresults}

\begin{figure}[t!]
     \centering
     \begin{subfigure}[b]{0.45\textwidth}
         \centering
         \includegraphics[width=\textwidth]{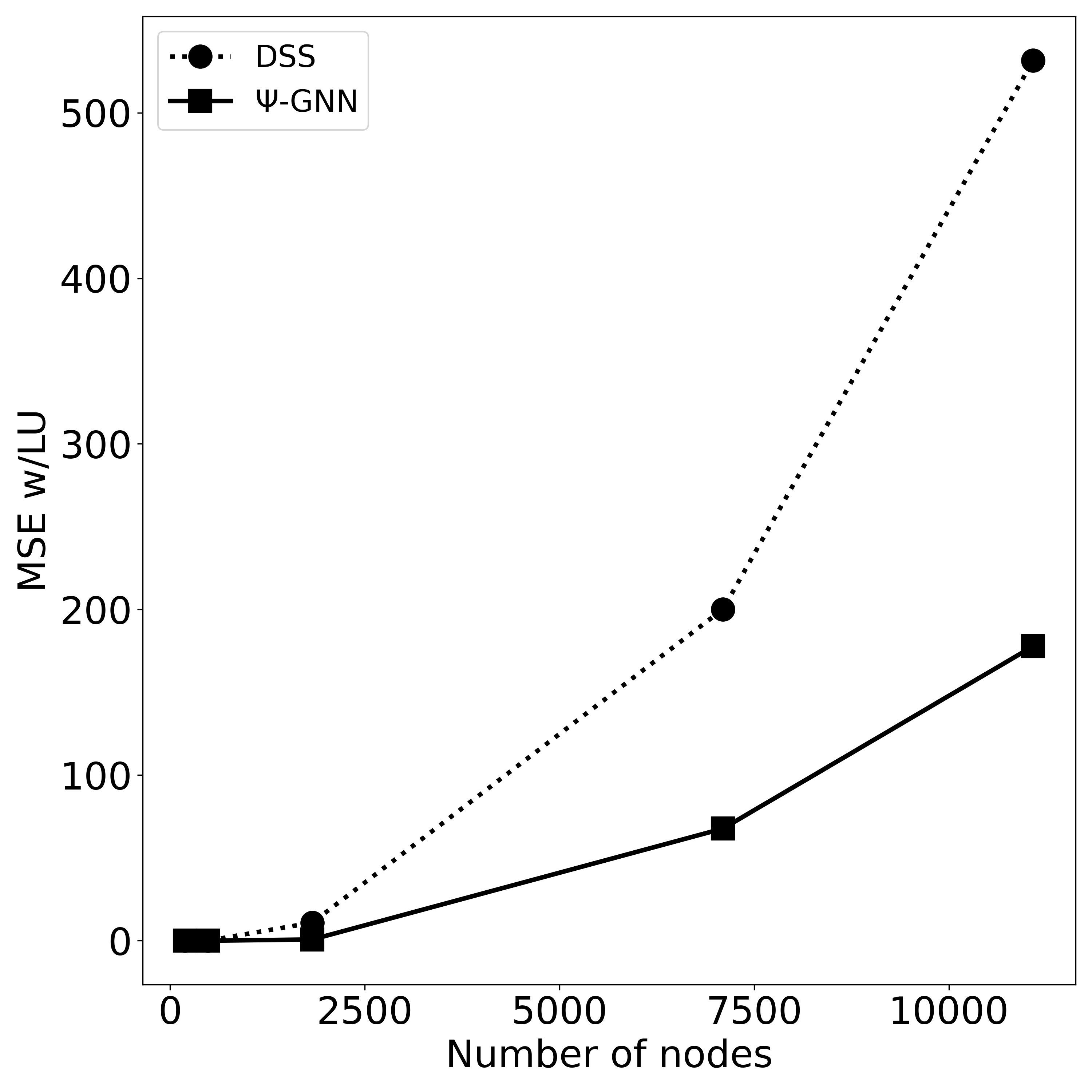}
         \caption{}
         \label{subfig:growing_geometries}
     \end{subfigure}
     \hfill
     \begin{subfigure}[b]{0.45\textwidth}
         \centering
         \includegraphics[width=\textwidth]{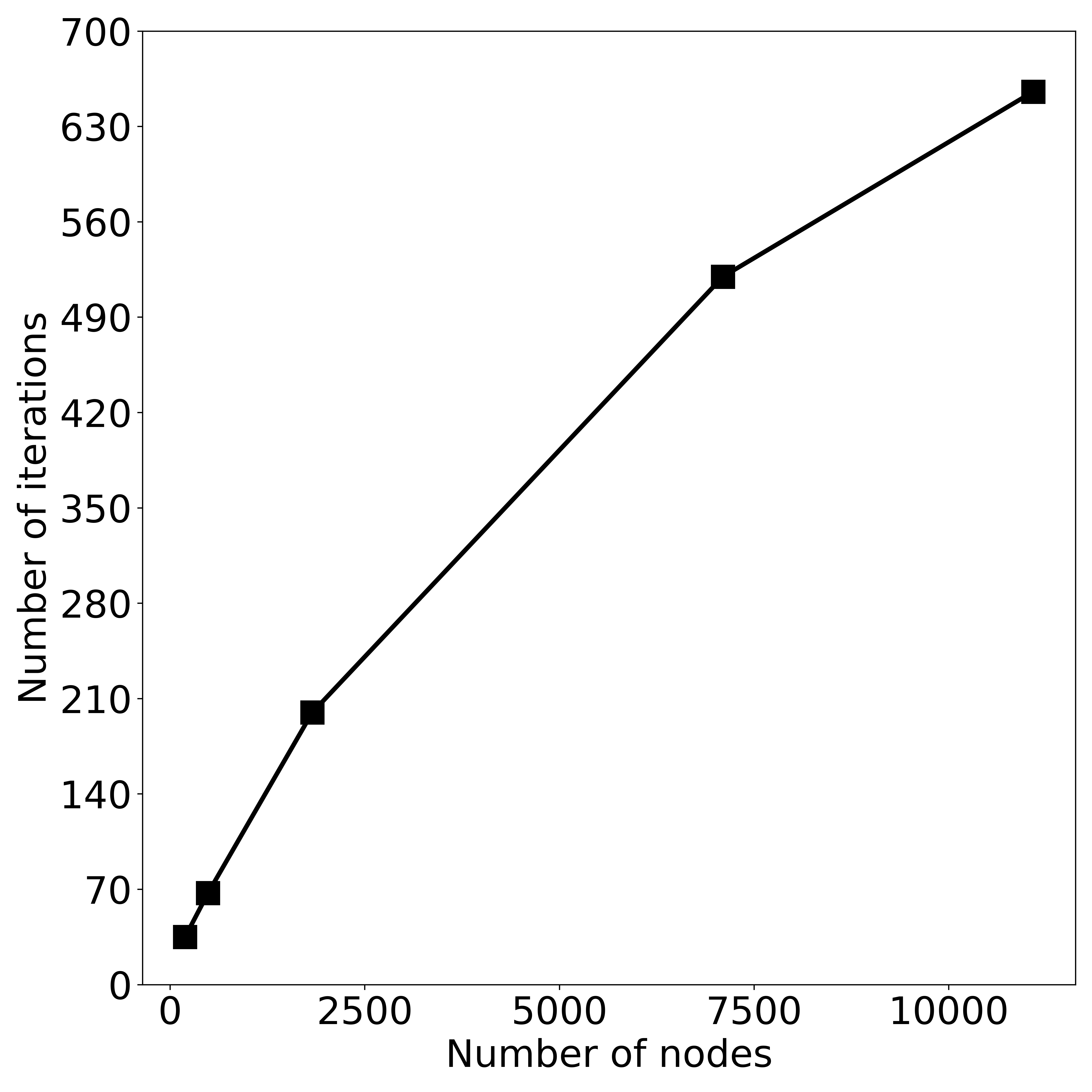}
         \caption{}
         \label{subfig:growing_iterations}
     \end{subfigure}
     \caption{Figure \ref{subfig:growing_geometries} illustrates the averaged MSE w/LU for different mesh sizes for DSS, and \name{}. Figure \ref{subfig:growing_iterations} displays the averaged number of iterations performed by the Broyden solver in \name{} to reach its target threshold with respect to the number of nodes per mesh.}
    \label{fig:sample_with_increasing_nodes}
\end{figure}

This section exhibits several generalizations aspects of \name{}. First, we demonstrate that \name{} remains consistent and outperforms DSS when solving problems with an increasing number of nodes in the graph. Then, we show the flexibility and insensitivity of \name{} with respect to any initial solution. Moreover, we provide evidence of the contractive nature of the constructed GNN function $h_\theta$, as discussed in Section \ref{stabilization}. Lastly, we address the resolution of an out-of-distribution problem, effectively highlighting the generalization capabilities of \name{}.
\\\\
\textbf{Size of the mesh} \quad This paragraph explores the performance and generalization capabilities of \name{} in solving Poisson problems with Dirichlet boundary conditions on meshes with a growing number of nodes, even though it was initially trained on meshes with approximately $500$ nodes. To conduct this experiment, we address the resolution of multiple Poisson problems on meshes with varying numbers of nodes. To maintain consistency in the distribution of inputs, we keep the same element sizes, similar to those used to generate the dataset, while varying the number of nodes by increasing the mesh radius. The force and boundary functions are randomly sampled following Section \ref{results:common_setup} but rescaled according to the selected radius. We consider six different setups corresponding to meshes with approximately $200$, $500$, $2000$, $7000$, and $11000$ nodes per mesh. For each setup, we solve $200$ Poisson problems using DSS and \name{}. The Broyden method used in the \name{} model is configured with a stopping criterion of $10^{-5}$ and a maximum of $1000$ iterations. The DSS model, as defined in Section \ref{results:direct_comparison}, infers solutions using only $30$ iterations. Notably, the \name{} model adjusts its iteration count thanks to the root-finding procedure, while DSS has a fixed number of iterations, and increasing them would necessitate training a new model. Figure \ref{subfig:growing_geometries} displays the MSE w/LU evolution (averaged over the $200$ problems) for each setup and each of the three models. The figure clearly shows that the DSS model diverges for larger meshes. In contrast, the \name{} model remains consistent when applied to larger meshes. Furthermore, Figure \ref{subfig:growing_iterations} displays the averaged iteration counts of the Broyden solver in the \name{} model with respect to the number of nodes per mesh. This figure clearly demonstrates that \name{} can adapt its number of Message-Passing layers to attain a solution.\\

\begin{figure}[t!]
    \centering
    \includegraphics[width=\textwidth]{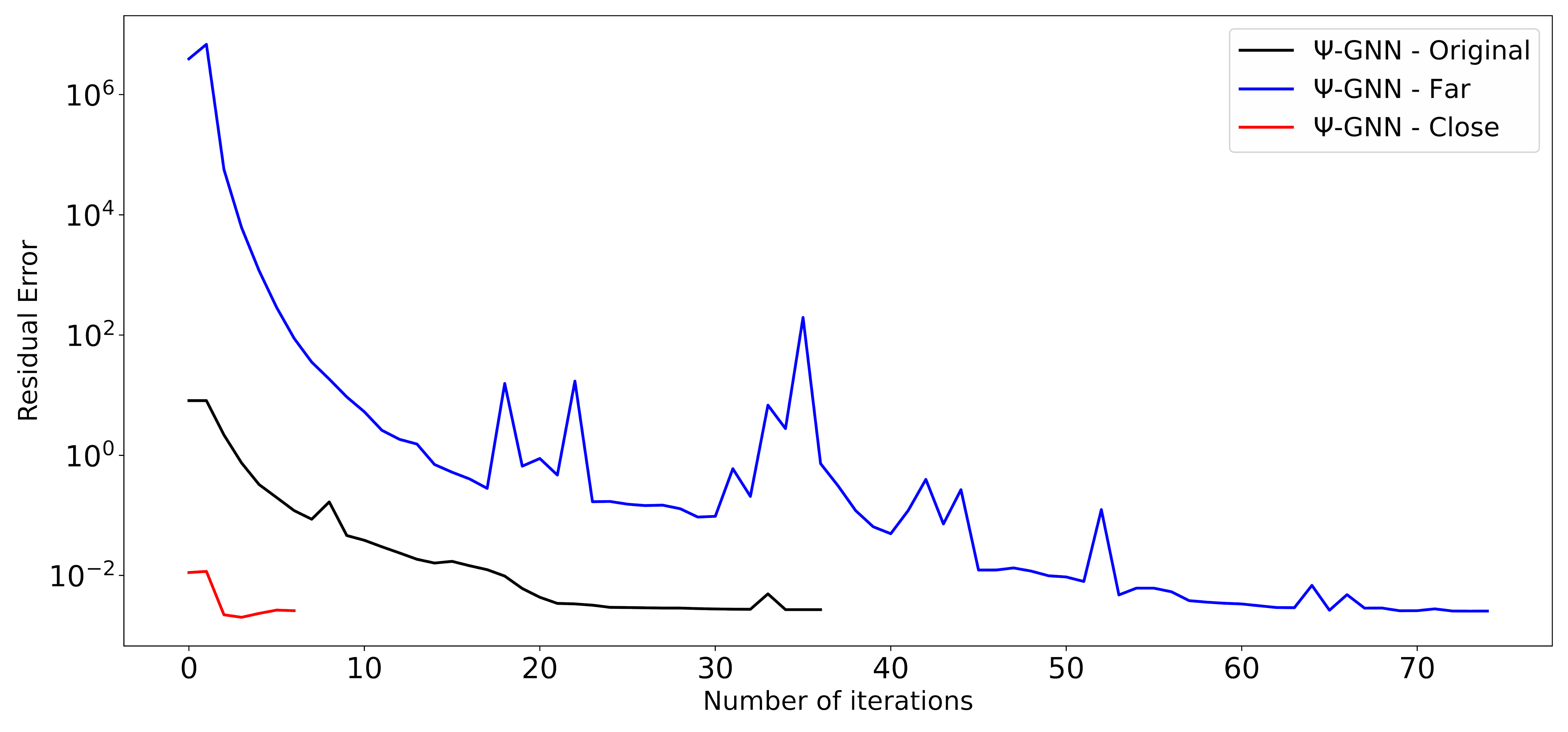}
    \caption{Evolution of the Residual error across iteration of \name{} for the same Poisson problem but considering three different initial solutions, demonstrating the adaptability of \name{} with respect to various initial solutions.}
    \label{fig:variousinit}
\end{figure}

\textbf{Initialization} \quad This paragraph aims to demonstrate the flexibility of \name{} and its insensitivity regarding its initialization. By ``flexible'', we mean that \name{} can dynamically adapt its number of iterations based on the distance from the initial solution to the final solution. And by ``insensitive'', we refer to the property that, regardless of the initial solution, the model consistently converges to the same fixed point, representing the desired solution. These advantageous characteristics are enabled by the auto-encoding process that maps the initial physical state to the latent space, where GNN layers are applied, and back. Flexibility is demonstrated thanks to Figure \ref{fig:variousinit}, which displays the evolution of the residual error across the iterations of the Broyden algorithm for the same Poisson problem considering three different initial conditions: the "Original" initial condition (black curve) is the one used during training, where the initial state is initialized to $0$. The "Far" initial condition (blue curve) involves applying a large random noise (uniform noise in the range $\left[-1000, 1000\right]$) to the solution obtained from the LU ``ground-truth" method. The "Close" condition (red curve) incorporates a small perturbation (uniform noise in the range $\left[0, 1\right]$) to the target ``ground-truth" solution. For all these initial conditions, the true value of Dirichlet boundary conditions is preserved. This experiment showcases that the number of iterations required for convergence varies depending on the proximity of the initial solution to the final solution. This adaptive behaviour enables the model to adjust its iteration count effectively, optimizing convergence efficiency. Insensitivity to initial solutions is demonstrated in the second column of Table \ref{tab:psignn_testsetneumann}, which evaluates the performance of \name{} using various initial solutions. This experiment demonstrates that \name{} is robust to the choice of initial solution: regardless of the initial solution provided, the algorithm consistently converges to the desired solution. Specifically, the "Random Initialization" entry in Table \ref{tab:psignn_testsetneumann} reports the metrics averaged over the entire test set, using an initial solution randomly generated by perturbing the ground-truth solution with uniformly random values ranging from $-1000$ to $1000$. Remarkably, the results obtained with random initialization are almost identical to those obtained with the ``Original setup'' (first column of \ref{tab:psignn_testsetneumann} identical to the results from Section \ref{results:mixed_bc}) indicating that our model is insensitive to the specific choice of the initial solution. In particular, this is a significant improvement compared to the DSS approach, which lacks this capability since it does not have any encoding process.\\

\begin{table}[t!]
\renewcommand{\arraystretch}{2.0}
\resizebox{\textwidth}{!}{%
\begin{tabular}{lccc}
% \cline{2-4}
\multicolumn{1}{c}{} & \textit{Original} &\textit{Random Initializer} & \textit{Forward Iteration} \\ \hline
\multicolumn{1}{l|}{Residual ($10^{-3}$)} & $3.16$ $\pm$ $0.2$& $3.18$ $\pm$ $0.3$ & $3.14$ $\pm$ $0.3$ \\ 
\hline
\multicolumn{1}{l|}{MSE w/LU} & $0.15$ $\pm$ $0.04$ & $0.16$ $\pm$ $0.09$ & $0.16$ $\pm$ $0.04$ \\ 
\hline
\end{tabular}%
}
\caption{Results averaged over the whole test set for different experiments. First column: original results of Section \ref{results:mixed_bc}; Second column: random initial solution; Third column: replacing Boyden with forward iterations.}
\label{tab:psignn_testsetneumann}
\end{table}

\textbf{Contractivity} \quad The regularization term (\ref{loss_frob}) serves the purpose of constraining the spectral radius of the Jacobian $J_{h_\theta}(\widehat{H})$ in order to ensure the stability of the model around the fixed point $\widehat{H}$, as discussed in Section \ref{stabilization}. 
The results obtained on the test set indicate the effectiveness of this regularization, with an average spectral radius of \textbf{0.993 $\pm$ 4e-4 $<$ 1}. This value is estimated a posteriori using the Power Iteration method \citep{power_iteration}. Furthermore, we demonstrate this generalization results in the third column of Table \ref{tab:psignn_testsetneumann}. In this experiment, we evaluate \name{} on the entire test set, but instead of using the root-finding Broyden algorithm employed during training, we simply iterate on the Processor described in the architecture in Section \ref{architecture}. Once again, the obtained results are almost identical to those obtained in the original setup. This demonstrates the contractive nature of the GNN-function $h_\theta$ and highlights the robustness and flexibility of \name{} in its ability to adapt to different solvers.\\

\begin{figure*}[t!]
\vskip 0.2in
\begin{center}
\centerline{\includegraphics[width=\textwidth]{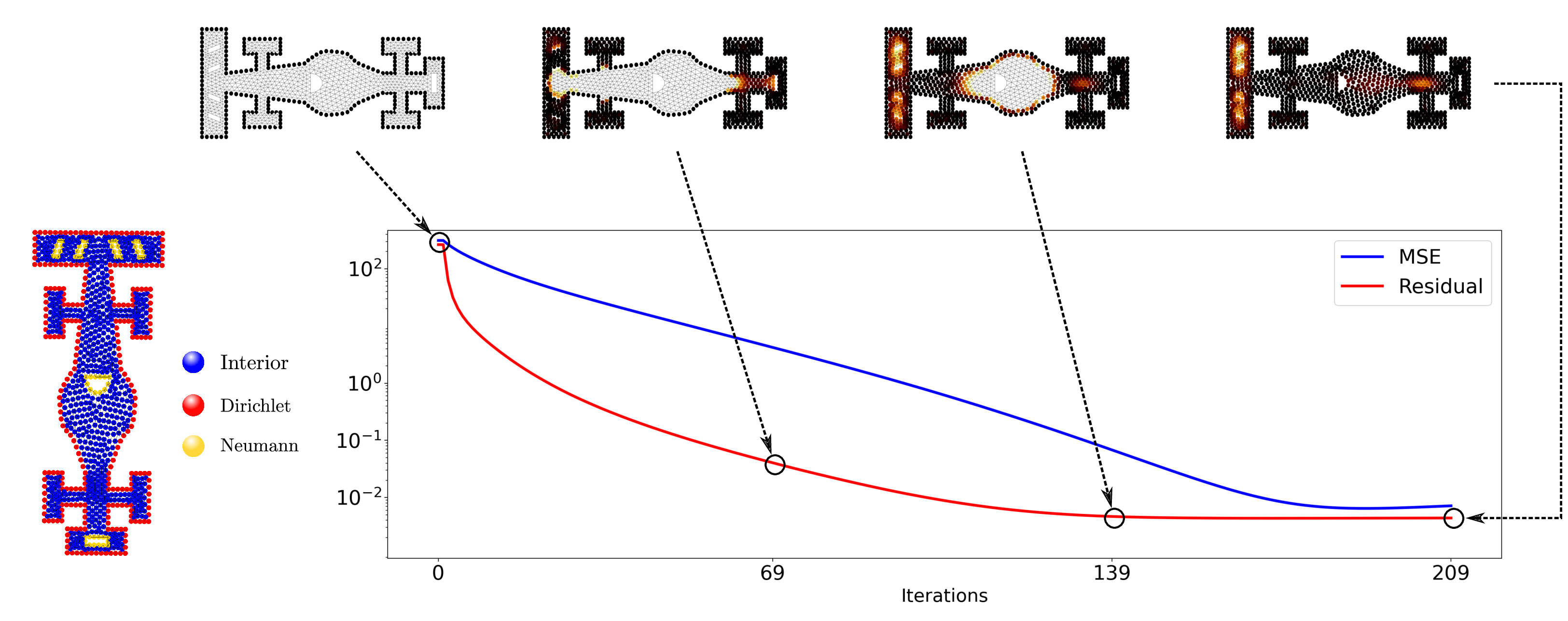}}
\caption{Generalization on the ``out-of-distribution" F1 shape, 1219 nodes. {\bf Central plot}: Residual and MSE during the 253 iterations of the Processor (i.e., without using RootFind), demonstrating the contractivity of $h_{\theta}$. {\bf Left}: The boundary conditions. {\bf Top}: the visual evolution of the squared error, displaying the flow of information from Dirichlet nodes inward.}
\label{img_evolution_f1}
\end{center}
\vskip -0.2in
\end{figure*}

\textbf{Out-of-distribution sample} \quad To further illustrate this, we conduct an experiment on a geometry representing a caricatural Formula 1 with 1219 nodes. This geometry includes ``holes" (such as a cockpit and front and rear wing stripes) and is larger (1219 nodes) and significantly different than those seen in the training dataset, providing a challenging test of the model's ability to generalize to out-of-distribution examples. We impose Dirichlet boundary conditions on all exterior nodes (pink nodes in the vertical plot at the left of Figure \ref{img_evolution_f1}) and Neumann conditions on the nodes within the ``holes" (yellow nodes). Functions $f$ and $g$ of (\ref{poisson-equation}) are randomly sampled from the same distribution as for the training set. The equilibrium of the model is found by simply iterating on the Processor instead of relying on Broyden's algorithm (see previous paragraph). The stopping criteria is the relative error set to $10^{-4}$. The results Figure \ref{img_evolution_f1} again shows the contracting nature of the learned function, that converges to the fixed point when iterated. Furthermore, it also gives an example of the generalization capacity of the learned model to some out-of-distribution examples. Additionally, the figure also illustrates how the information propagates through the graph, starting from the Dirichlet nodes to gradually filling the whole domain.

\subsection{Inference complexity}
\label{subsec:complexity}

To determine the expected performance of \name{} as the number of nodes $N$ in a graph varies, it is essential to analyze its complexity. In \name{}, all neural networks have a single hidden layer of dimension $d$, so the complexity of applying a neural network to one node is $\mathcal{O}\left(d^3\right)$. Let us assume that $m$ is the average number of neighbours for each node in the graph. The complexity of computing the output of a GNN layer relative to one node in a mesh is then of $\mathcal{O}\left(md^3\right)$. Considering all $N$ nodes in the graph and iterating for $K$ updates, the complexity becomes $\mathcal{O}\left(KNmd^3\right)$. This represents the theoretical complexity of \name{} when the Processor is iterated upon for $K$ iterations. \\

When using \name{} with Broyden's algorithm, the complexity is found to be of a higher order. Broyden's method is a quasi-Newton method that computes the next iterate $H^{k+1}$ as follows:

\begin{equation}
    H^{k+1} = H^k + J^{-1}_{|M_\theta}(H^k)M_\theta(H^k)
    \label{eq:broyden1}
\end{equation}

In Equation \eqref{eq:broyden1}, we already know the complexity of $M_\theta(H^k)$, which is computed in a similar manner as previously explained and has a complexity of $\mathcal{O}\left(Nmd^3\right)$. In his paper \citep{broyden}, Broyden suggests using the Sherman-Morrison formula \citep{sherman-morrison} to update the inverse of the Jacobian matrix directly, as follows:

\begin{equation}
    B_{|M_\theta}^{k} = B_{|M_\theta}^{k-1} + \dfrac{\Delta H^k - B_{|M_\theta}^{k-1}\Delta M_\theta^k}{(\Delta H^k)^TB_{|M_\theta}^{k-1}\Delta M_\theta^k}(\Delta H^k)^TB_{|M_\theta}^{k-1}
    \label{eq:sherman_morrison}
\end{equation}

Here, $B_{|M_\theta}^{k} = J^{-1}_{|M_\theta}(H^k)$, $\Delta H^k = H^k - H^{k-1}$, and $\Delta M_\theta^k = M_\theta(H^k) - M_\theta(H^{k-1})$. 

In Equation \eqref{eq:sherman_morrison}, all operations are matrix-vector products of size $N$, so the complexity of updating the Jacobian matrix is $\mathcal{O}\left(N^2\right)$. Back to equation \eqref{eq:broyden1}, the total complexity to compute the next iterate is then of $\mathcal{O}\left(N^2\right) + \mathcal{O}\left(Nmd^3\right)$, which corresponds to the cost of updating the Jacobian matrix and the cost of evaluating the GNN model. This is applied for $M$ iterations of the Broyden algorithm, resulting in a global complexity of $\mathcal{O}\left(MN^2\right)$, since the quadratic complexity dominates the linear one.

\section{Conclusions and Future Work}
\label{conclusion}

This paper has introduced \name{}, a novel and consistent Machine Learning based approach that combines Graph Neural Networks and Implicit Layer Theory to effectively solve a wide range of Poisson problems. The model, trained in a ``physics-informed" manner, is found to be robust, stable, and highly adaptable to varying mesh sizes, domain shapes, boundary conditions, and initialization. To the best of our knowledge, this approach is distinct from any previous Machine Learning based methods. Furthermore, \name{} can be extended to other steady-state partial differential equations and its application to 3-dimensional domains is straightforward. Overall, the results of this study demonstrate the significant potential of \name{} in solving Poisson problems in a consistent and efficient manner.
\\\\
The long-term objective of the program driving this work is to accelerate industrial Computation Fluid Dynamics (CFD) codes on software platforms such as OpenFOAM \cite{openfoam}. In future work, we aim to evaluate the efficiency of the proposed approach by implementing it in an industrial CFD code, and assessing its impact on performance acceleration. However, despite its flexibility, the proposed method faces limitations in terms of graph size. Scaling up the model to handle industrial-scale cases poses a significant challenge that \name{} alone, in its current state, cannot overcome. This challenge arises due to the increasing number of iterations required and the limited expressiveness of GNN models when applied to large graphs. Nonetheless, there are promising directions to scale up the model. One approach is to employ it as a solver for domain decomposition algorithms, leveraging the batch parallel framework of Machine Learning models to solve multiple sub-problems simultaneously \citep{nastorg2024multi}. Another potential direction for future research is the exploration of a fully hierarchical architecture.

%%%%%%%%%%%%%%%%%%%%%%%%%%%%%%%%%%%%%%%%%%%%%%%%%%%%%%%%%%%%%%%%%%%%%%%%%%%%%%%%%%%%%%%%%%
\newpage 

\section*{Acknowledgment}
This research was supported by DATAIA Convergence Institute as part of the ``Programme d’Investissement d’Avenir'', (ANR- 17-CONV-0003) operated by INRIA and IFPEN.\\

During the preparation of this work, the authors used OpenAI ChatGPT in order to rectify grammatical errors. After using this tool, the authors reviewed and edited the content as needed and take full responsibility for the content of the publication.

\bibliographystyle{elsarticle-harv}
\bibliography{the_biblio}

%%%%%%%%%%%%%%%%%%%%%%%%%%%%%%%%%%%%%%%%%%%%%%%%%%%%%%%%%%%%%%%%%%%%%%%%%%%%%%%%%%%%%%%%%%

\newpage

\appendix

\section{Brief introduction to the Finite Element method}
\label{appendix:fem}

This section provides a brief introduction to the fundamental concepts of the Finite Element method (FEM), explaining how to construct the discrete system described in \eqref{linear-system} from the considered Poisson problem (\ref{poisson-equation}).
\\\\
Problem (\ref{poisson-equation}) consists in solving the following 2d boundary-value problem:
\begin{equation}
\left \{
\begin{array}{rcl}
-\Delta u &=& f \qquad \in \Omega \\
u &=& g \qquad \in \partial \Omega_D \\
\frac{\partial u}{\partial n} &=& 0 \qquad \in \partial \Omega_N
\end{array}
\right.
\label{app:fem1}
\end{equation}
where $u=u(x,y)$ is the unknown function, $f = f(x,y)$ is the forcing function and $g = g(x,y)$ is the Dirichlet boundary function defined on the 2d domain $\Omega$ with boundary $\partial \Omega = \Omega_D \cup \Omega_N$. Besides, $n$ denotes the outward normal vector on $\partial \Omega$ and $\Delta$ is the Laplace operator defined as: 
\begin{equation*}
    \Delta u = \dfrac{\partial^2 u}{\partial^2 x} + \dfrac{\partial^2 u}{\partial^2 y}
\end{equation*}
The FEM usually consists of the following steps: i): conversion of the PDE into a variational formulation ii): discretization of the domain $\Omega$ into a mesh $\Omega_h$ iii): choice of interpolation functions iv): formulation of the discretized system v): resolution of the system.
\\\\
The variational form of the PDE is obtained by multiplying the PDE by a function $v$ and integrating it over the domain $\Omega$. The function $v$ is known as the test function, while the solution function $u$ is referred to as the trial function. Both test and trial functions belong to some suitable function spaces denoted as $V$ for the trial function space and $\widehat{V}$ for the test function space. Specifically, $V$ requires that $u=g$ on $\partial\Omega_D$ while $\widehat{V}$ requires that $v$ vanishes on $\partial \Omega_D$. Using Green's formula, this setting yields the following variational problem (for more advanced theoretical details, refer to \citet{intro_num_meth_var_prob}):
\\\\
\textit{Find} $u \in V$ \textit{such that:} 
\begin{equation}
    \int_\Omega \nabla u \cdot \nabla v \dx = \int_\Omega fv \dx \qquad \forall v \in \widehat{V}
    \label{app:fem4}
\end{equation}
The variational formulation (\ref{app:fem4}) poses a continuous problem that defines the solution $u$ in an infinite-dimensional space $V$. In contrast, the FEM pursues an approximate solution $u_h$ of $u$ by substituting the infinite-dimensional function spaces with finite ones. To that purpose, the continuous domain $\Omega$ is first discretized into an unstructured triangular mesh $\Omega_h$ with $N$ nodes. Next, let $V_h$ be a finite approximation space constructed using first-order piecewise polynomials defined on each triangle $K$ of $\Omega_h$. Furthermore, let ($\phi_i)_{(1\leq i\leq N)}$ be a set of basis functions for $V_h$. Notably, these basis functions are constructed to satisfy the following property:
\begin{equation*}
\phi_i(x_j) = \delta_{i,j} \quad \text{with} \quad \delta_{i,j} =   \left \{
\begin{array}{rcl}
&1 &\text{if } i = j \\
&0 &\text{otherwise} 
\end{array}
\right.
\label{app:propertybasisfunctions}
\end{equation*} 
One can consider an extension of $u_h$ on the basis of $V_h$ such that:
\begin{equation}
    u_h = \sum_{i=1}^N u_i\phi_i
    \label{app:approx}
\end{equation}
Taking $v_i = \phi_i$, the variational form \ref{app:fem4} can be discretized such that: 
\begin{align}
\sum_{j=1}^N u_j  \int_\Omega \nabla \phi_j \cdot \nabla \phi_i \dx  = \int_\Omega f_i \phi_i \dx  \qquad \forall ~ 1\leq i\leq N
\label{app:fem6}
\end{align}
Introducing the \textit{stiffness} matrix
\begin{equation*}
    A = (a_{ij})_{(1 \leq i,j \leq N)} \quad \text{with} \quad a_{ij} = \int_\Omega \nabla \phi_j \cdot \nabla \phi_i \dx 
\end{equation*}
the solution vector $U = (u_i)_{(1\leq i\leq N)}$ and the right-hand side vector
\begin{equation*}
    B = (b_i)_{(1\leq i\leq N)} \quad \text{with} \quad b_i = \int_\Omega f_i \phi_i ~ \text{dx}
\end{equation*}
then (\ref{app:fem6}) corresponds to a linear system to solve of the form:
\begin{equation}
    AU = B
    \label{app:linearsystem}
\end{equation}
It is important to note that using this specific discretization approach does not impose any boundary conditions. In the case of homogeneous Neumann boundary conditions, the variational formulation \ref{app:fem4} suggests that these conditions should not be considered during the discretization process, which is consistent with the previous configuration. However, the treatment of Dirichlet boundary conditions is not handled automatically. Instead, in practice, the Dirichlet boundary conditions are incorporated by modifying the linear system (\ref{app:linearsystem}) manually: for the indices corresponding to Dirichlet boundary conditions, the corresponding row of matrix $A$ is set to 0, and a 1 is placed on the diagonal. Additionally, the value in vector $B$ at the corresponding index is changed to match the discrete value of $g$.
\\\\
To further illustrate this, let's consider the resolution of the following 1-d Poisson problem with mixed boundary conditions, defined on $\Omega = (0,1)$: 
\begin{equation}
\left \{
\begin{array}{rcl}
-u''(x) &=& f(x) \qquad x \in (0,1) \\
u(0) &=& g_0  \\
u'(1) &=& 0
\end{array}
\right.
\end{equation}
In this example, we assume that $\Omega$ is divided into 4 cells with equal lengths, denoted as $h = \frac{1}{4}$. Using the previously described approach, we arrive at the following linear system \ref{app:linearsystem}:
\begin{equation}
\underbrace{
\frac{1}{h}\begin{bmatrix*}
    1 & -1 & 0 & 0 & 0 \\
    -1 & 2 & -1 & 0 & 0 \\
    0 & -1 & 2 & -1 & 0 \\
    0 & 0 & -1 & 2 & -1 \\
    0 & 0 & 0 & -1 & 1 \\
\end{bmatrix*}}_A
\underbrace{
\begin{bmatrix*}
    u_0 \\
    u_1 \\
    u_2 \\
    u_3 \\
    u_4 \\
\end{bmatrix*}}_{U}
=
\underbrace{
\begin{bmatrix*}
    hf_0 \\
    hf_1 \\
    hf_2 \\
    hf_3 \\
    hf_4 \\
\end{bmatrix*}}_{B}  
\label{app:matrix1}
\end{equation}
As mentioned previously, solving \ref{app:matrix1} does not fulfil Dirichlet boundary conditions (i.e. $u_0 \neq g_0$). Manually changing the matrix to impose these conditions yields the following system:

\begin{equation*}
\underbrace{
\frac{1}{h}\begin{bmatrix*}
    \textcolor{red}{h} & \textcolor{red}{0} & \textcolor{red}{0} & \textcolor{red}{0} & \textcolor{red}{0} \\
    -1 & 2 & -1 & 0 & 0 \\
    0 & -1 & 2 & -1 & 0 \\
    0 & 0 & -1 & 2 & -1 \\
    0 & 0 & 0 & -1 & 1 \\
\end{bmatrix*}}_{A}
\underbrace{
\begin{bmatrix*}
    u_0 \\
    u_1 \\
    u_2 \\
    u_3 \\
    u_4 \\
\end{bmatrix*}}_{U}
=
\underbrace{
\begin{bmatrix*}
    \textcolor{red}{g_0} \\
    hf_1 \\
    hf_2 \\
    hf_3 \\
    hf_4 \\
\end{bmatrix*}}_{B} 
\end{equation*}
Considering the sparse matrix $A$ as an adjacency matrix for its corresponding mesh, it is possible to analyse the structure of the induced graph. The first row of $A$ implies $u_0 = g_0$, which indicates that $u_0$ remains independent of information from other nodes. However, the second row of $A$ results in:
\begin{equation}
    u_1 = \dfrac{h}{2}\left(\frac{1}{h}u_0 + \frac{1}{h}u_2 + hf_1\right)
    \label{app:fem7}
\end{equation}
\eqref{app:fem7} implies that computing $u_1$ requires information from both $u_0$ and $u_2$, resulting in an undirected graph for this particular node. By applying the same analysis, one can deduce that matrix $A$ represents a directed graph at Dirichlet nodes, where information is only sent, and an undirected graph for Interior (rows 2 to 4) and Neumann (row 5) nodes, where information is both sent and received.

\newpage 

\section{Supplementary materials for training}
\label{supplementary_materials_training}

This Appendix provides additional training materials detailing the architecture of some trainable functions and explaining the format of vectors considered in \ref{architecture}. It also informs of how the normalization is performed on the data.
\\\\

\textbf{Construction of $\Lambda_{i,\theta}$} \quad Equation \eqref{resnetfashion} details the process of updating an Interior node. As a reminder, the updated Interior latent state $z_{i}^\texttt{I}$ is computed such that:
\begin{align}
z_{i}^\texttt{I} &= H_i + \Lambda_{i,\theta}
\label{app:resnet}
\end{align}
To compute $\Lambda_{i,\theta}$, we use two MLPs, $\Psi^1_\theta$ and $\Psi^2_\theta$, which both take the same inputs. These inputs include the current latent state $H_i$, computed MPNNs (\ref{mp_int_in}) and (\ref{mp_int_out}), and problem-specific data $b_i$ (refer to the details below). The MLPs compute $\alpha_i$ and $\zeta_i$, respectively. The final $\Lambda_{i,\theta}$ is obtained by multiplying $\alpha_i$ and $\zeta_i$ as follows:
\begin{align}
\alpha_i &= \Psi^1_\theta(H_i, b_i, \phi_{\rightarrow,i}^\texttt{I}, \phi_{\leftarrow,i}^\texttt{I})
\label{alpha}\\
\zeta_i &= \Psi^2_\theta(H_i, b_i, \phi_{\rightarrow,i}^\texttt{I}, \phi_{\leftarrow,i}^\texttt{I})
\label{zeta} \\
\Lambda_{i,\theta} &= \alpha_i \times \zeta_i
\end{align}
The key distinction lies in the construction of $\Psi^1_\theta$ and $\Psi^2_\theta$. Although they share the same architecture as described in the ``General experimental setup'' from Section \ref{results:common_setup} (i.e. one hidden layer with dimension $10$), $\Psi^1_\theta$ uses a Sigmoid activation function, while $\Psi^2_\theta$ uses a ReLU activation function. As a result, $\alpha_i$ is restricted to $\left(0,1\right)$, preventing the updating process (\ref{app:resnet}) from exploding. Furthermore, $\alpha_i$ can be interpreted as an attention layer, prioritizing the most crucial connections to enhance the performance of our model.
\\\\
\textbf{Data information} \quad The architecture of the model described in \ref{architecture} relies on several inputs whose format needs to be precise. \ref{alpha}, \ref{zeta} and \ref{update_neumann},  require as input problem-related data, encoded into a vector $b_i$ such that:
\begin{equation}
    b_i = \left \{
    \begin{array}{r c l}
        \begin{bmatrix*}[r]
            f_i & 0 & 0
        \end{bmatrix*}  & \quad & i \text{ is Interior} \\ 
        \begin{bmatrix*}[r]
             0 & g_i & 0
        \end{bmatrix*} &\quad & i \text{ is Dirichlet} \\
        \begin{bmatrix*}[r]
            0 & 0 & f_i
        \end{bmatrix*} &\quad & i \text{ is Neumann} 
    \end{array}
    \right.
    \label{b-vector}
\end{equation}  
where $f_i$ and $g_i$ are the discretized values of the volumetric function $f$ and the Dirichlet boundary function $g$ from the Poisson problem (\ref{poisson-equation}).
\\\\
\textbf{Data normalization} \quad In such problems, normalizing the input features is essential to enhance the model's performance during the training phase. To achieve this, the distances $d_{ij}$ in equations (\ref{mp_int_in}), (\ref{mp_int_out}), (\ref{mp_neu_in}), the problem-related vector $b_i$ in equations (\ref{alpha}), (\ref{zeta}), and (\ref{update_neumann}), as well as the normal vector $n_i$ in equation (\ref{update_neumann}), are all normalized. This normalization is accomplished by subtracting the mean and dividing by the standard deviation, which are computed based on the entire dataset. When dealing with problem-related data, the normalization is conducted column-wise, taking into account the statistics of the entire dataset. This approach is necessary as $f_i$ and $g_i$ may represent a multi-modal distribution, rendering it challenging to achieve proper normalization if treated as a whole. \textbf{Consequently, all inputs used for solution inference are normalized.}

\newpage 

\section{Implicit Models}
\label{implicit_models}
This section delves into the Implicit Layer Theory, providing supplementary information on its implementation. The motivations for its use are outlined in \ref{motivations_ilt}, followed by a detailed description of the training procedures in \ref{training_ilt}, emphasising backpropagation and exploiting Theorem 1 in \citet{deq}. Furthermore, a comprehensive analysis of the Hutchinson method \citep{hutchinson} used to calculate the regularizing term as defined equation (\ref{loss_frob}) is presented in \ref{regularizing_term_ilt}.

\subsection{Motivations}
\label{motivations_ilt}
\begin{figure}[!ht]
     \centering
     \begin{subfigure}[b]{1.\textwidth}
         \centering
         \includegraphics[scale=0.8]{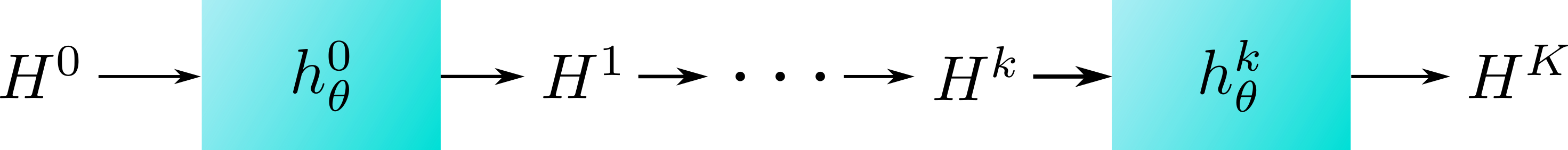}
         \vspace{0.3cm}
         \caption{Stacking GNN layers : $H^{k+1} = h_\theta^k\left(H^k\right)$}
         \label{fig:archidss}
     \end{subfigure}
     \par\bigskip\bigskip
     \begin{subfigure}[b]{1.\textwidth}
         \centering
         \includegraphics[scale=0.9]{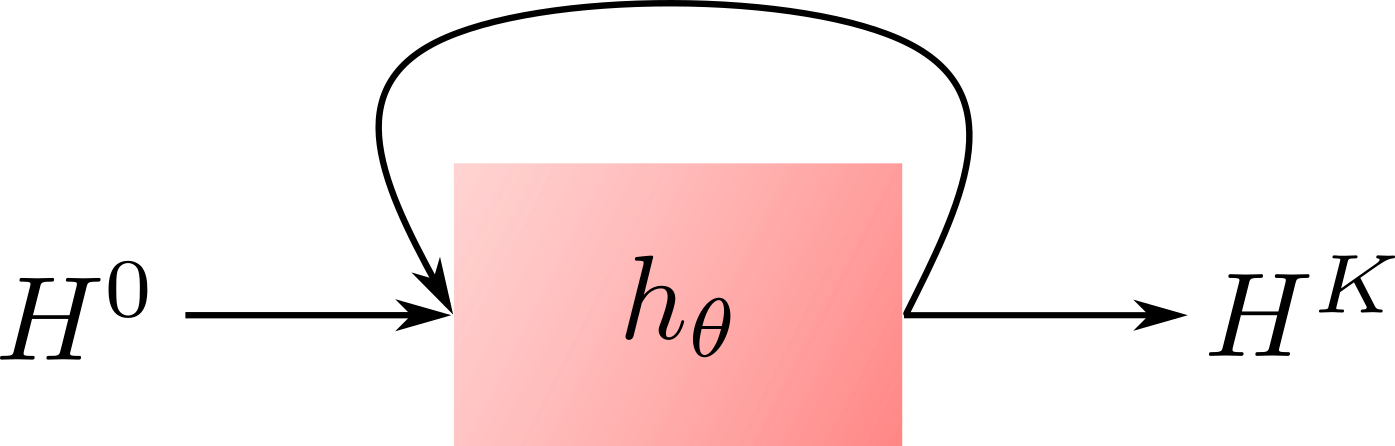}
         \vspace{0.3cm}
         \caption{Recurrent GNN : $H^{k+1} = h_\theta\left(H^k\right)$}
         \label{fig:archirnn}
     \end{subfigure}
     \par\bigskip\bigskip
     \begin{subfigure}[b]{1.\textwidth}
         \centering
         \includegraphics[scale=0.9]{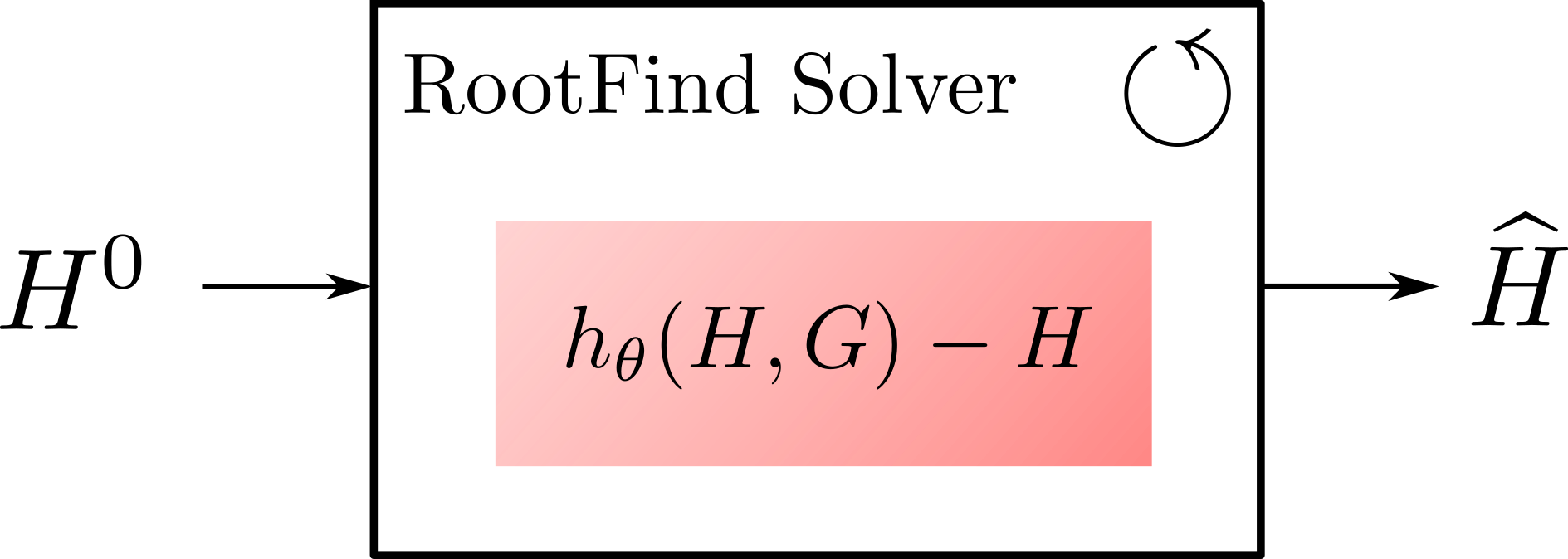}
         \vspace{0.3cm}
         \caption{\name{} : $\widehat{H} = \text{RootFind}\left( h_\theta(H,G) - H \right)$}
         \label{fig:archipsignn}
     \end{subfigure}
     \caption{Example of three types of GNN architecture. Figure \ref{fig:archidss} displays the DSS approach, which consists of stacking different GNN layers until the final state is reached. Figure \ref{fig:archirnn} represents the Recurrent-GNN method, iterating over a GNN function $h_\theta$ for a fixed number $K$ of iterations the final state is reached. Figure \ref{fig:archipsignn} illustrates the proposed \name{} approach. The final solution $\widehat{H}$ is computed as the equilibrium point of the $h_\theta$ function using a black-box RootFind solver. The solver is initialized with an initial condition $H^0$, and the Message Passing iterations are performed implicitly within the solver.}
     \label{fig:architecturetypes}
\end{figure}

Many architectures are available for training Graph Neural Network (GNN) models, as demonstrated in \citet{survey_graph_network}. In the context of this work, where we aim to solve a Poisson problem on unstructured meshes by directly minimizing the residual of the discretized equation, the number of layers required to achieve convergence should be proportional to the diameters of the meshes under consideration. Figure \ref{fig:archidss} and \ref{fig:archirnn} illustrate the two common GNN architectures traditionally used for this purpose. In Figure \ref{fig:archidss}, the final solution $H^K$ is obtained after iterating a fixed number of times $K$ on different Message Passing layers $h^i_\theta$, each with distinct weights. This architecture is employed in approaches such as in \citet{deep_statistical_solver}. In Figure \ref{fig:archirnn}, the architecture is of recurrent type, where the final solution is obtained by also iterating a fixed number of times $K$ but on the same neural network $h_\theta$. This architecture has the advantage of significantly reducing the size of the model, as employed in the work of \citet{ds_gps}. In both methods, the number of iterations $K$ (i.e. the number of Message Passing steps) is fixed, limiting the model's ability to generalize to different mesh sizes. \textbf{However, using the recurrent architecture, it has been experimentally demonstrated that the model, trained with a sufficient number of iterations, tends towards a fixed point.} \\

% \begin{figure}[ht]
% \vskip 0.2in
% \begin{center}
% % \centerline{\includegraphics[width=\textwidth]{img_appendix/motivationRootFind.png}}
% \centerline{\includesvg[width=\textwidth]{img_dss_recurrent_archi.svg}}
% \caption{Comparison between a usual GNN architecture (top) where different Message Passing layers $h_\theta^i$ are stacked (i.e. the weights are different for all layers) and a Recurrent-GNN structure (bottom) where the solution $\widehat{H}$ is computed iterating on the same Message Passing layer.}
% \label{img_motivationRootFind}
% \end{center}
% \vskip -0.2in
% \end{figure}

Building upon these previous experimental results, we propose to enhance the recurrent architecture by incorporating a black-box root-finding solver to directly find the equilibrium point of the model, as illustrated in figure \ref{fig:archipsignn}. In this approach, the iterations (i.e. the flow of information) are performed implicitly within the RootFind solver. This method eliminates the need for a fixed number of iterations, as the number of required Message Passing steps is now exclusively determined by the precision imposed on the solver, resulting in a more adaptable and flexible approach.

\subsection{Training an Implicit Model}
\label{training_ilt}

Following \citet{deq}, the training of an implicit model requires, at each iteration, the resolution of two fixed point problems: one for the forward pass and one for the backward pass. 
\\\\
\textbf{Forward pass} \quad The resolution of a fixed point for the forward pass is clear and represents the core of our approach. A root-finding solver (i.e. a quasi-Newton method to avoid computing the inverse Jacobian of a Newton method at each iteration step) is used to determine the fixed point $\widehat{H}$ of the function $h_\theta$ (\ref{eq_func}) such that: 
\begin{align}
    \widehat{H} & = \text{RootFind}\left( h_\theta(H,G) - H \right)
\end{align}
\textbf{Backward pass} \quad However, using a black-box solver prevents the use of explicit backpropagation through the exact operations performed in the forward pass. Thankfully, \citet{deq} proposes a simpler alternative procedure that requires no knowledge of the black-box solver by directly computing the gradient at the fixed point. The gradient of the loss $\mathcal{L}$ with respect to the weights $\theta$ is then given by: 
\begin{equation}
    \frac{\partial \mathcal{L}}{\partial \theta} = - \frac{\partial \mathcal{L}}{\partial \widehat{H}}\left(J_{h_\theta}^{-1}|_{\widehat{H}}\right)\frac{\partial h_\theta(\widehat{H}, G)}{\partial \theta}.
    \label{loss_backward}
\end{equation}
where $\left(J_{h_\theta}^{-1}|_{\widehat{H}}\right)$ is the inverse Jacobian of $h_\theta$ evaluated at $\widehat{H}$. To avoid computing the expensive $- \frac{\partial \mathcal{L}}{\partial \widehat{H}}\left(J_{h_\theta}^{-1}|_{\widehat{H}}\right)$ term in (\ref{loss_backward}), one can alternatively solve the following root finding problem using Broyden's method (or any other root-finding solver) and the autograd packages from Pytorch:
\begin{equation}
    \left(J_{h_\theta}^T|_{\widehat{H}}\right)x^T + \left(\frac{\partial \mathcal{L}}{\partial \widehat{H}}\right)^T = 0
    \label{backward_linear_system}
\end{equation}
Consequently, the model is trained using a RootFind solver to compute the linear system (\ref{backward_linear_system}) and directly backpropagate through the equilibrium using (\ref{loss_backward}).

\subsection{Computing the regularizing term}
\label{regularizing_term_ilt}

In Section \ref{stabilization}, we propose to rely on the method outlined in \citet{deq_stab} to regularize the model's conditioning. The spectral radius of the Jacobian $\rho\left(J_{h_\theta}\right)$, responsible for the stability of the model around the fixed point $\widehat{H}$, is constrained by directly minimizing its Frobenius norm since it is an upper bound for the spectral radius. This method prevents the use of computationally expensive methods (i.e. Power Iteration \cite{power_iteration} for instance). The Frobenius norm is estimated using the classical Hutchinson estimator (\cite{hutchinson}):
\begin{equation}
||J_{h_\theta}||^2_F = \mathbb{E}_{\epsilon \in \mathcal{N}(0,I_d)}\left[||\epsilon^T J_{h_\theta}||^2_2\right]
\label{frob_norm}
\end{equation}
where $J_{h_\theta} \in \mathbb{R}^{d\times d}$.
The expectation (\ref{frob_norm}) can be estimated using a Monte-Carlo method for which (empirically observed) a single sample suffices to work well.

%%%%%%%%%%%%%%%%%%%%%%%%%%%%%%%%%%%%%%%%%%%%%%%%%%%%%%%%%%%%%%%%%%%%%%%%%%%%%%%%%%%%%%%%%%

\newpage 

\section{Theoretical proofs}
\label{appendix:theory}

\begin{pf}[Proposition \ref{prop:equivalence}]\label{pf:equivalence}\mbox{}\\
If $h^\star_G$ is a solution to the problem \eqref{eq:psignn_optiproblemfixedpoint}, then its fixed point is a candidate solution to the problem \eqref{eq:psignn_optiproblem} and we have:

\begin{equation*}
\mathcal{L}_\text{res}(\text{FixedPoint}(h^\star_G),~G) \geqslant \mathcal{L}_\text{res}(U^\star_G,~G)    
\end{equation*}

Reciprocally, for a fixed $G$, if $U^\star_G$  is a solution to the problem \eqref{eq:psignn_optiproblem}, then the function $h_G(H) = U^\star_G$ which always outputs the same value has a unique fixed point, namely $U^\star_G$. Considering this function as a candidate to the problem \eqref{eq:psignn_optiproblemfixedpoint}, we have: 

\begin{align*}
\mathcal{L}_\text{res}(U_G^\star,~G) = \mathcal{L}_\text{res}(\text{FixedPoint}\left(h_G\right),~G) \geqslant \mathcal{L}_\text{res}(\text{FixedPoint}(h^\star_G),~G)   
\end{align*}

Consequently, 

\begin{equation*}
\mathcal{L}_\text{res}(U_G^\star,~G) = \mathcal{L}_\text{res}(\text{FixedPoint}(h^\star_G),~G)    
\end{equation*}

and the problems are equivalent.
\end{pf}

\begin{pf}[Proposition \ref{prop:satisfyconditions}]\label{pf:satisfyconditions}\mbox{}\\
Corollary $1$ in \citet{deep_statistical_solver} assumes that four hypotheses must be fulfilled:

\begin{enumerate}
\item The loss function $\mathcal{L}_\text{res}$ is continuous and permutation-invariant.
\item The solution $U^\star_G$ is unique.
\item The problem distribution $G$ satisfies permutation-invariance, compactness, connectivity (each graph having a single connected component), and separability of external outputs (ensuring node identifiability).
\item The solution is continuous with respect to $G$.
\end{enumerate}

The first hypothesis is immediately satisfied by the specific loss function, which is suitable for problems involving GNNs and similar to the one described in \citet{deep_statistical_solver}. Additionally, the existence and uniqueness of the solution are guaranteed thanks to a FEM analysis of problem \eqref{poisson-equation} which includes at least one Dirichlet boundary condition. This analysis, based on the Lax-Milgram theorem \citep{fem_larson}, validates the second hypothesis. Regarding the properties of the problem distribution, our dataset generator (see Section \ref{results:common_setup}) generates graphs with smooth boundaries within a bounded domain, following a rotation-equivariant law and ensuring that its inside has only one connected component. Node identifiability within a graph is achieved through the use of edge features representing distances to the closest neighbors, which are never equal, thereby validating the third hypothesis. If full identifiability is desired (not just almost surely), an additional descriptor can be added to each node. Experiments have actually been run in that setting, with no observable difference in performance, thereby validating the third hypothesis. The continuity of the solution $U^\star_G$ with respect to $G$ depends on the specific choice of the loss function $\mathcal{L}_\text{res}$. For Poisson-like problems similar to those considered in this study, continuity can be established based on the following Lemma \ref{pf:continuity}, and this will conclude the proof.
\end{pf}

\begin{lem}[Continuity of $\varphi$]\label{pf:continuity}\mbox{}\\
The mapping
\begin{equation*}
\varphi: \;\;\; 
\begin{array}{rcl}
\mathcal{S} & \rightarrow & \mathbb{R}^N \\ G = (N, A, B) &\;\mapsto\; & U^\star(G) :=\underset{U}{\argmin}~\| AU-B\|^2    
\end{array}
\end{equation*}
is continuous w.r.t.~$A$ and $B$.
\end{lem}

\begin{pf}[Lemma \ref{pf:continuity}]\mbox{}\\
Linear systems such as \eqref{linear-system}, which result from the FEM discretization of the Poisson problem \eqref{poisson-equation}, have a unique solution given by $U^\star(G) = A^{-1}B$. This solution is linear in $B$ and thus continuous with respect to $B$. The question at hand is whether the mapping from $A$ to its inverse $A^{-1}$ is also continuous. We can express $A^{-1}$ as

\begin{equation*}
A^{-1} = \frac{\text{adj}(A)}{\text{det}(A)}   
\end{equation*}

where $\text{adj}(A)$ denotes the adjoint of $A$ and $\text{det}(A)$ represents the determinant of $A$. Both the adjoint operation and the determinant are continuous operations. The remaining aspect to consider is whether the determinant of $A$ can be zero or not. In the specific settings where $A$ arises from the discretization of \eqref{poisson-equation}, including boundary conditions, the determinant of $A$ is always non-negative. This completes the proof.
\end{pf}

\begin{pf}[Theorem \ref{thm:universal_approx}]\label{pf:theorem}\mbox{}\\
The architecture of $\widehat\varphi$ obtained by Corollary \ref{cor:exsitence_gnn} can be decomposed into two blocks: a $\widehat\varphi_{\text{GNN}}$ which consists of GNN layers that are applied to a hidden state $H$, and a $\widehat\varphi_{\text{Dec}}$ block that represents the decoder (see Section \ref{architecture}). Here, we attempt to build a function whose fixed point w.r.t. $H$ approximates the output of the  $\widehat\varphi_{\text{GNN}}$ block. To do this, we consider the function

\begin{equation*}
h_{\theta_\varepsilon}:~~~~
\begin{array}{rcl}
\mathbb{R}^{N \times d} \times \mathcal{S} &\rightarrow& \mathbb{R}^{N \times d} \\ 
(H,~G) &\mapsto& \widehat\varphi_{\text{GNN}}(G)
\end{array}
\end{equation*}

where $d$ is the dimension of the hidden space. For any fixed $G$, this function $h_{\theta_\varepsilon}$ is constant w.r.t.~$H$, and, consequently has a unique fixed point w.r.t.~$H$, which is $\widehat\varphi_{\text{GNN}}(G)$. As \name{} encompasses the $\widehat \varphi$ architecture, we can indeed represent exactly $h_{\theta_\varepsilon}$ using the \name{} Processor defined in Section \ref{architecture}. The complete architecture of \name{} can be written as follows:

\begin{equation*}
\widehat U = \text{Dec}\left(\text{FixedPoint}(h_{\theta_\varepsilon}\left(\text{Enc}\left(U\right), G\right)\right) 
\end{equation*}

where the decoder $\text{Dec}$ is $\widehat\varphi_{\text{Dec}}$. The encoder $\text{Enc}$, which is an additional feature of \name{} architecture, can be chosen arbitrarily since, in this proof, $h_{\theta_\varepsilon}$ always outputs the same value, regardless of $H$. This completes the proof.
\end{pf}

\begin{pf}[Proposition \ref{prop:contractivity}]\label{pf:contractivity}\mbox{}\\
The trainable function $h_{\theta_\varepsilon}(H,G)$ from Theorem \ref{thm:universal_approx} can be assumed to be an approximation of a contractive function $f$ (e.g., as built in the proof). For the sake of simplicity, we omit the Decoder here. Thus, there exists $\lambda < 1$ such that, for any problem $G$ and any latent values $H$, $H'$:

\begin{equation*}
d( f(H), f(H')) \;\leqslant\; \lambda\, d(H,H')    
\end{equation*}

where $d$ denotes the Euclidean distance. Note that $\lambda \in [0,1[$ can even be $0$, e.g. for $f(H) = U^\star_G$. Consequently, we have :

\begin{equation*}
d( h(H), h(H')) \;\leqslant\; d(f(H), f(H')) + 2\varepsilon \;\leqslant\; \lambda\, d(H,H') + 2\varepsilon   
\end{equation*}

given that $\|h(H)  - f(H)\| \leqslant \varepsilon$ and $\|h(H')  - f(H')\| \leqslant \varepsilon$. Suppose $\varepsilon$ is small enough such that $\mu = \lambda + 2\sqrt{\varepsilon} < 1$. Then, for $H$, $H'$ satisfying $d(H,H') > \sqrt{\varepsilon}$, we have:

\begin{equation*}
d( h(H), h(H')) \;\leqslant\; \mu \,d(H,H')   
\end{equation*}

since $\lambda\, d(H,H') + 2\varepsilon < \mu\, d(H,H')$ is equivalent to $d(H,H') > \frac{2\varepsilon}{\mu-\lambda} = \sqrt{\varepsilon}$. Thus $h$ is contractive for all pairs of points farther than  $\sqrt{\varepsilon}$ from each other. In particular, if $d(H,h^\star_G) > \sqrt{\varepsilon}$, then $d( h(H), h(h^\star_G)) \leqslant \mu \,d(H,h^\star_G)$, which implies the exponential convergence of an iterative power method from any initialization $H$ to the ball of radius $\sqrt{\varepsilon}$ around the fixed point $h^\star_G$, which is itself at distance at most $\varepsilon$ from the true optimal solution $U^\star_G$. Thus the function $h$ is predominantly contractive in that sense. Using $\varepsilon' = \varepsilon + \sqrt{\varepsilon}$ then gets rid of square roots.  
\end{pf}

\end{document}